\documentclass[10pt,twocolumn,letterpaper]{article}

\usepackage{cvpr}
\usepackage{times}
\usepackage{epsfig}
\usepackage{graphicx}
\usepackage{amsmath}
\usepackage{amssymb}
\usepackage{multirow}
\usepackage{graphicx}
\usepackage{subfigure}
\usepackage{color}
\usepackage{booktabs}
\usepackage{tabulary}
\usepackage{rotating}
\usepackage{bbding}

\newcommand{\bx} {{\bf x }}

\newcommand{\by} {{\bf y }}

\usepackage[pagebackref=true,breaklinks=true,letterpaper=true,colorlinks,bookmarks=false]{hyperref}

\cvprfinalcopy 

\def\cvprPaperID{875} 

\ifcvprfinal\fi
\begin{document}

\title{WIDER FACE: A Face Detection Benchmark}

\author{Shuo Yang\\\and Ping Luo\\\and Chen Change Loy\\\and Xiaoou Tang\\\and
\small{Department of Information Engineering, The Chinese University of Hong Kong}\\
{\tt\small \{ys014, pluo, ccloy, xtang\}@ie.cuhk,edu.hk}
}

\maketitle

\begin{abstract}
\label{sec:abstract}
Face detection is one of the most studied topics in the computer vision community.
Much of the progresses have been made by the availability of face detection benchmark datasets. We show that there is a gap between current face detection performance and the real world requirements.
To facilitate future face detection research, we introduce the WIDER FACE dataset, which is $10$ times larger than existing datasets. 
The dataset contains rich annotations, including occlusions, poses, event categories, and face bounding boxes. Faces in the proposed dataset are extremely challenging due to large variations in scale, pose and occlusion, as shown in Fig.~\ref{fig:wider_intro}. Furthermore, we show that WIDER FACE dataset is an effective training source for face detection.
We benchmark several representative detection systems, providing an overview of state-of-the-art performance and propose a solution to deal with large scale variation.  
Finally, we discuss common failure cases that worth to be further investigated. Dataset can be downloaded at: \url{mmlab.ie.cuhk.edu.hk/projects/WIDERFace}

\end{abstract}

\section{Introduction}
\label{sec:introduction}
\begin{figure}[t]
\begin{center}
\includegraphics[width=1\linewidth]{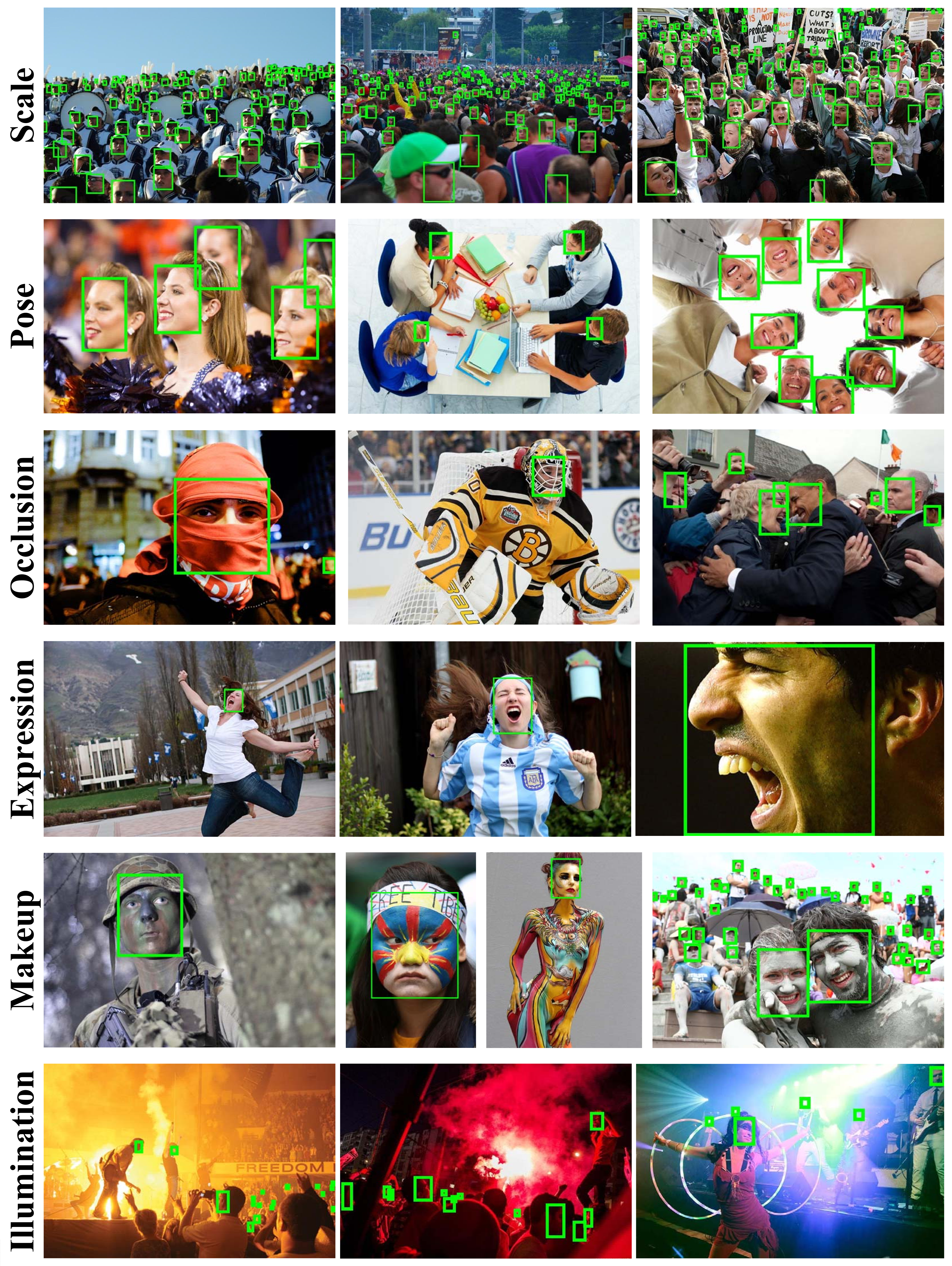}
\vspace {-0.25cm}
\caption{\small{ We propose a WIDER FACE dataset for face detection, which has a high degree of variability in scale, pose, occlusion, expression, appearance and illumination. We show example images (cropped) and annotations. The annotated face bounding box is denoted in green color. The WIDER FACE dataset consists of $393,703$ labeled face bounding boxes in $32,203$ images \bf{(Best view in color)}.}}
\vskip -0.5cm
\label{fig:wider_intro}
\end{center}
\end{figure}

Face detection is a critical step to all facial analysis algorithms, including face alignment~\cite{zhangalignment15,lbf,sdm}, face recognition~\cite{deepface}, face verification~\cite{celeface,facenet}, and face parsing~\cite{face_parsing}.
Given an arbitrary image, the goal of face detection is to determine whether or not there are any faces in the image and, if present, return the image location and extent of each face~\cite{Yang02survey}.
While this appears as an effortless task for human, it is a very difficult task for computers.
The challenges associated with face detection can be attributed to variations in pose, scale, facial expression, occlusion, and lighting condition, as shown in Fig.~\ref{fig:wider_intro}.
Face detection has made significant progress after the seminal work by Viola and Jones~\cite{viola2004robust}.
Modern face detectors can easily detect near frontal faces and are widely used in real world applications, such as digital camera and electronic photo album.
Recent research~\cite{JointCascade, cascadecnn, HeadHunter, ACF-multiscale, yang2015faceness} in this area focuses on the unconstrained scenario, where a number of intricate factors such as extreme pose, exaggerated expressions, and large portion of occlusion can lead to large visual variations in face appearance.

Publicly available benchmarks such as FDDB~\cite{fddb}, AFW~\cite{zhu2012face}, PASCAL FACE~\cite{yan2014face}, have contributed to spurring interest and progress in face detection research. However, as algorithm performance improves, more challenging datasets are needed to trigger progress and to inspire novel ideas. Current face detection datasets typically contain a few thousand faces, with limited variations in pose, scale, facial expression, occlusion, and background clutters, making it difficult to assess for real world performance. As we will demonstrate, the limitations of datasets have partially contributed to the failure of some algorithms in coping with heavy occlusion, small scale, and atypical pose.

In this work, we make three contributions. (1) We introduce a large-scale face detection dataset called WIDER FACE. It consists of $32,203$ images with $393,703$ labeled faces, which is $10$ times larger than the current largest face detection dataset~\cite{ijb15}. The faces vary largely in appearance, pose, and scale, as shown in Fig.~\ref{fig:wider_intro}. In order to quantify different types of errors, we annotate multiple attributes: occlusion, pose, and event categories, which allows in depth analysis of existing algorithms. (2) We show an example of using WIDER FACE through proposing a multi-scale two-stage cascade framework, which uses divide and conquer strategy to deal with large scale variations. Within this framework, a set of convolutional networks with various size of input are trained to deal with faces with a specific range of scale. (3) We benchmark four representative algorithms~\cite{HeadHunter, viola2004robust, ACF-multiscale, yang2015faceness}, either obtained directly from the original authors or reimplemented using open-source codes. We evaluate these algorithms on different settings and analyze conditions in which existing methods fail.

\section{Related Work}
\label{sec:related work}

\noindent
\textbf{Brief review of recent face detection methods:}
Face detection has been studied for decades in the computer vision literature. Modern face detection algorithms can be categorized into four categories: cascade based methods~\cite{JointCascade,huang2007high,SURF,npdface,viola2004robust}, part based methods~\cite{dp2mfd,yan2014face,zhu2012face}, channel feature based methods~\cite{yang2015ccf,ACF-multiscale}, and neural network based methods~\cite{ddfd,cascadecnn,yang2015faceness}. Here we highlight a few notable studies. A detailed survey can be found in~\cite{Yang02survey,zhang2010survey}.
The seminal work by Viola and Jones~\cite{viola2004robust} introduces integral image to compute Haar-like features in constant time. These features are then used to learn AdaBoost classifier with cascade structure for face detection.
Various later studies follow a similar pipeline. Among those variants, SURF cascade~\cite{SURF} achieves competitive performance. Chen~\etal~\cite{JointCascade} learns face detection and alignment jointly in the same cascade framework and obtains promising detection performance. 

One of the well-known part based methods is deformable part models (DPM)~\cite{dpm}. Deformable part models define face as a collection of parts and model the connections of parts through Latent Support Vector Machine. The part based methods are more robust to occlusion compared with cascade-based methods.
A recent study~\cite{HeadHunter} demonstrates state-of-the art performance with just a vanilla DPM, achieving better results than more sophisticated DPM variants~\cite{yan2014face,zhu2012face}.
Aggregated channel feature (ACF) is first proposed by Dollar~\etal~\cite{dollar09acf} to solve pedestrian detection. Later on, Yang~\etal~\cite{ACF-multiscale} applied this idea on face detection. In particular, features such as gradient histogram, integral histogram, and color channels are combined and used to learn boosting classifier with cascade structure.
Recent studies~\cite{cascadecnn, yang2015faceness} show that face detection can be further improved by using deep learning, leveraging the high capacity of deep convolutional networks. 
We anticipate that the new WIDER FACE data can benefit deep convolutional network that typically requires large amount of data for training.

\vspace{0.1cm}
\noindent
\textbf{Existing datasets:}
We summarize some of the well-known face detection datasets in Table~\ref{tab:cmp_dataset}.
AFW~\cite{zhu2012face}, FDDB~\cite{fddb}, and PASCAL FACE~\cite{yan2014face} datasets are most widely used in face detection. AFW dataset is built using Flickr images. It has $205$ images with $473$ labeled faces. For each face, annotations include a rectangular bounding box, $6$ landmarks and the pose angles. FDDB dataset contains the annotations for $5,171$ faces in a set of $2,845$ images. PASCAL FACE consists of $851$ images and $1,341$ annotated faces. Recently, IJB-A~\cite{ijb15} is proposed for face detection and face recognition. IJB-A contains $24,327$ images and $49,759$ faces. MALF is the first face detection dataset that supports fine-gained evaluation. MALF~\cite{yanfacedataset15} consists of $5,250$ images and $11,931$ faces.
The FDDB dataset has helped driving recent advances in face detection. However, it is collected from the Yahoo! news website which biases toward celebrity faces. The AFW and PASCAL FACE datasets contain only a few hundred images and has limited variations in face appearance and background clutters. The IJB-A dataset has large quantity of labeled data; however, occlusion and pose are not annotated. The MAFL dataset labels fine-grained face attributes such as occlusion, pose and expression. The number of images and faces are relatively small.
Due to the limited variations in existing datasets, the performance of recent face detection algorithms saturates on current face detection benchmarks. For instance, on AFW, the best performance is $97.2\%$ AP; on FDDB, the highest recall is $91.74\%$; on PASCAL FACE, the best result is $92.11\%$ AP. The best few algorithms have only marginal difference. 
\begin{table}[t]
\begin{center}
\caption{Comparison of face detection datasets.}
\label{tab:cmp_dataset}
\vskip 0.25cm
\tiny\addtolength{\tabcolsep}{-1pt}
\addtolength{\tabcolsep}{-1pt}
\begin{tabular}{ c  || c | c || c | c || c | c | c || c | c | c}
\hline
& \multicolumn{2}{c||}{Training}
& \multicolumn{2}{c||}{Testing}
& \multicolumn{3}{c||}{Height}
& \multicolumn{3}{c}{Properties} \\ \cline{2-11}

\textbf{Dataset} & \rotatebox{90}{\#Image} & \rotatebox{90}{\#Face} & \rotatebox{90}{\#Image} & \rotatebox{90}{\#Face} & \rotatebox{90}{10-50 pixels}& \rotatebox{90}{50-300 pixels} & \rotatebox{90}{$\leq$300 pixels} & \rotatebox{90}{Occlusion labels} & \rotatebox{90}{Event labels} & \rotatebox{90}{Pose labels} \\

\hline\hline
AFW~\cite{zhu2012face}& - & - & 0.2k & 0.47k & 12\% & 70\% & 18\% & - & - & \Checkmark\\
FDDB~\cite{fddb}& - & - & 2.8k & 5.1k & 8\% & 86\% & 6\% & - & - & -\\
PASCAL FACE~\cite{yan2014face}& - & - & 0.85k & 1.3k & 41\% & 57\% & 2\% & - & - & -\\
IJB-A~\cite{ijb15}& 16k  & 33k & 8.3k & 17k & 13\% & 69\% & 18\% & - & - & -\\
MALF~\cite{yanfacedataset15}& - & - & 5.25k & 11.9k & N/A & N/A & N/A & \Checkmark & - & \Checkmark\\
\textbf{WIDER FACE}& 16k & \textbf{199}k & 16k &\textbf{194}k & 50\% & 43\% & 7\% & \Checkmark & \Checkmark & \Checkmark\\
\hline
\end{tabular}
\end{center}
\end{table}

\section{WIDER FACE Dataset}
\label{sec:dataset}
\subsection{Overview}
\label{sec:overview}
To our knowledge, WIDER FACE dataset is currently the largest face detection dataset, of which images are selected from the publicly available WIDER dataset~\cite{xiongevent15}. We choose $32,203$ images and label $393,703$ faces with a high degree of variability in scale, pose and occlusion as depicted in Fig.~\ref{fig:wider_intro}. WIDER FACE dataset is organized based on $60$ event classes. For each event class, we randomly select $40\%$/$10\%$/$50\%$ data as training, validation and testing sets. Here, we specify two training/testing scenarios:
\begin{itemize}
\item\textbf{Scenario-Ext:} A face detector is trained using any external data, and tested on the WIDER FACE test partition.  
\item\textbf{Scenario-Int:} A face detector is trained using WIDER FACE training/validation partitions, and tested on WIDER FACE test partition.
\end{itemize}
We adopt the same evaluation metric employed in the PASCAL VOC dataset~\cite{pascalvoc}.  
Similar to MALF~\cite{yanfacedataset15} and Caltech~\cite{dollarcaltech09} datasets, we do not release bounding box ground truth for the test images. Users are required to submit final prediction files, which we shall proceed to evaluate.
     
\subsection{Data Collection}
\label{sec:data collection}
\begin{figure}[t]
\begin{center}
\includegraphics[width=1\linewidth]{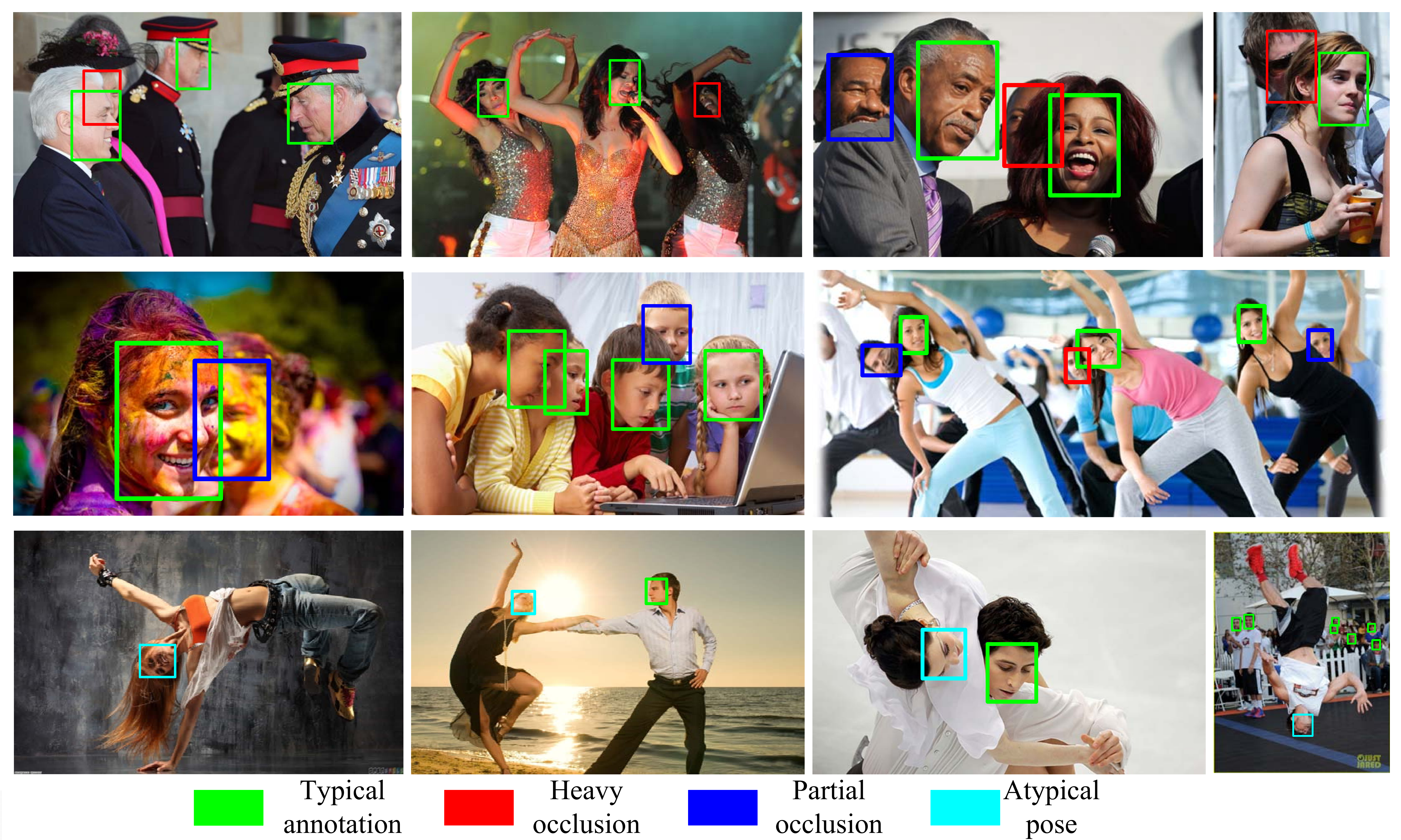}
\caption{\small{Examples of annotation in WIDER FACE dataset \bf{(Best view in color)}.}}
\label{fig:wider_anno}
\end{center}
\vspace{-0.5cm}
\end{figure}

\noindent\textbf{Collection methodology}. WIDER FACE dataset is a subset of the WIDER dataset~\cite{xiongevent15}. The images in WIDER were collected in the following three steps: 1) Event categories were defined and chosen following the Large Scale Ontology for Multimedia (LSCOM)~\cite{lscom06}, which provides around $1,000$ concepts relevant to video event analysis. 2) Images are retrieved using search engines like \emph{Google} and \emph{Bing}. For each category, $1,000$-$3,000$ images were collected. 3) The data were cleaned by manually examining all the images and filtering out images without human face. Then, similar images in each event category were removed to ensure large diversity in face appearance. A total of $32,203$ images are eventually included in the WIDER FACE dataset.\\
\noindent\textbf{Annotation policy}. We label the bounding boxes for all the recognizable faces in the WIDER FACE dataset. The bounding box is required to tightly contain the forehead, chin, and cheek, as shown in Fig.~\ref{fig:wider_anno}. If a face is occluded, we still label it with a bounding box but with an estimation on the scale of occlusion. Similar to the PASCAL VOC dataset~\cite{pascalvoc}, we assign an 'Ignore' flag to the face which is very difficult to be recognized due to low resolution and small scale ($10$ pixels or less). After annotating the face bounding boxes, we further annotate the following attributes: pose (typical, atypical) and occlusion level (partial, heavy). Each annotation is labeled by one annotator and cross-checked by two different people.   

\subsection{Properties of WIDER FACE}
\label{sec:properties}

\begin{figure}[t]
\begin{center}
\includegraphics[width=1\linewidth]{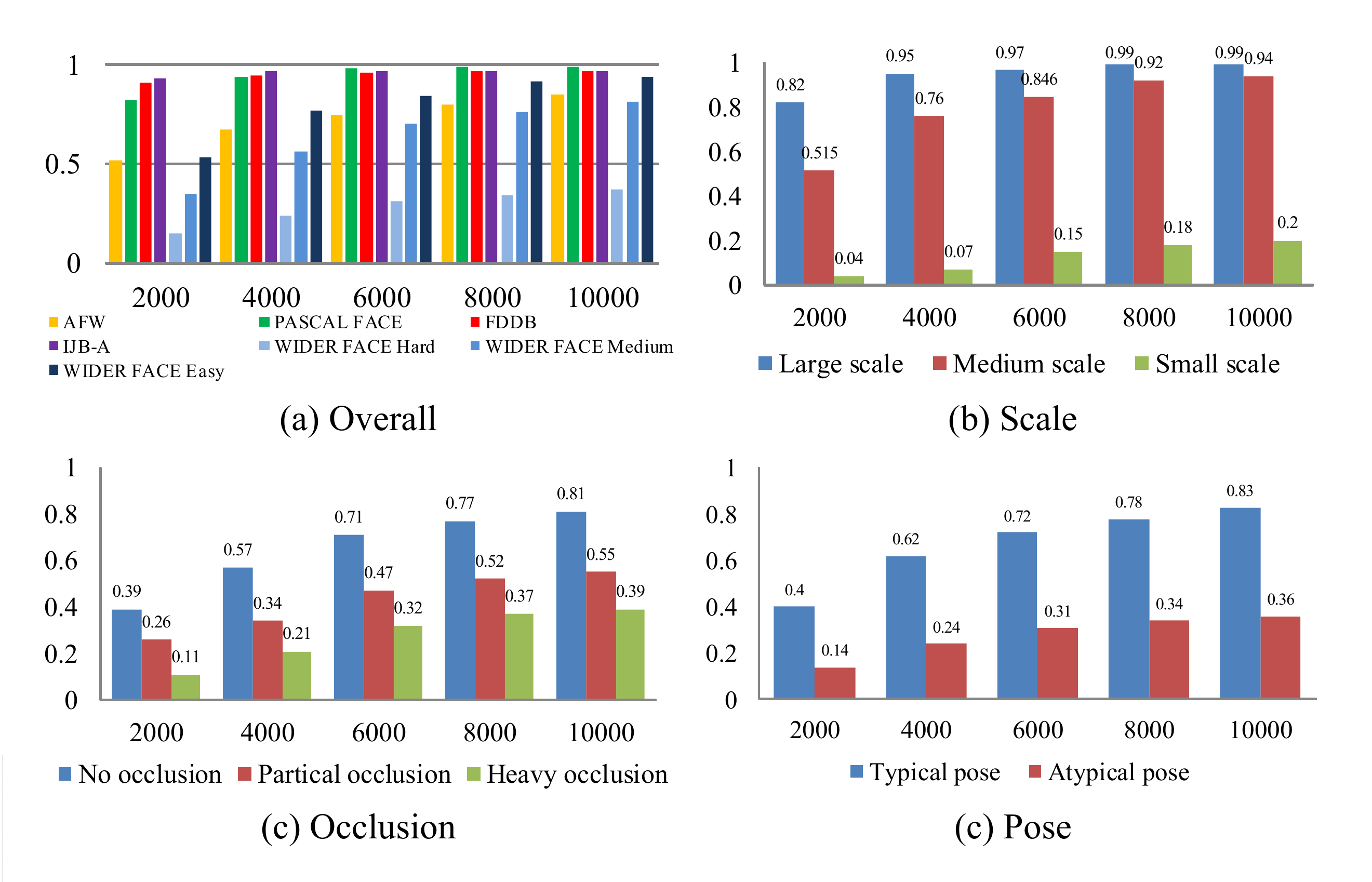}
\vskip -0.25cm
\caption{\small{The detection rate with different number of proposals. The proposals are generated by using Edgebox~\cite{edgebox}. Y-axis denotes for detection rate. X-axis denotes for average number of proposals per image. Lower detection rate implies higher difficulty. We show histograms of detection rate over the number of proposal for different settings (a) Different face detection datasets. (b) Face scale level. (c) Occlusion level. (d) Pose level.}}
\vskip -0.5cm
\label{fig:wider_stats}
\end{center}
\end{figure}

\begin{figure*}[t]
\begin{center}
\includegraphics[width=\linewidth]{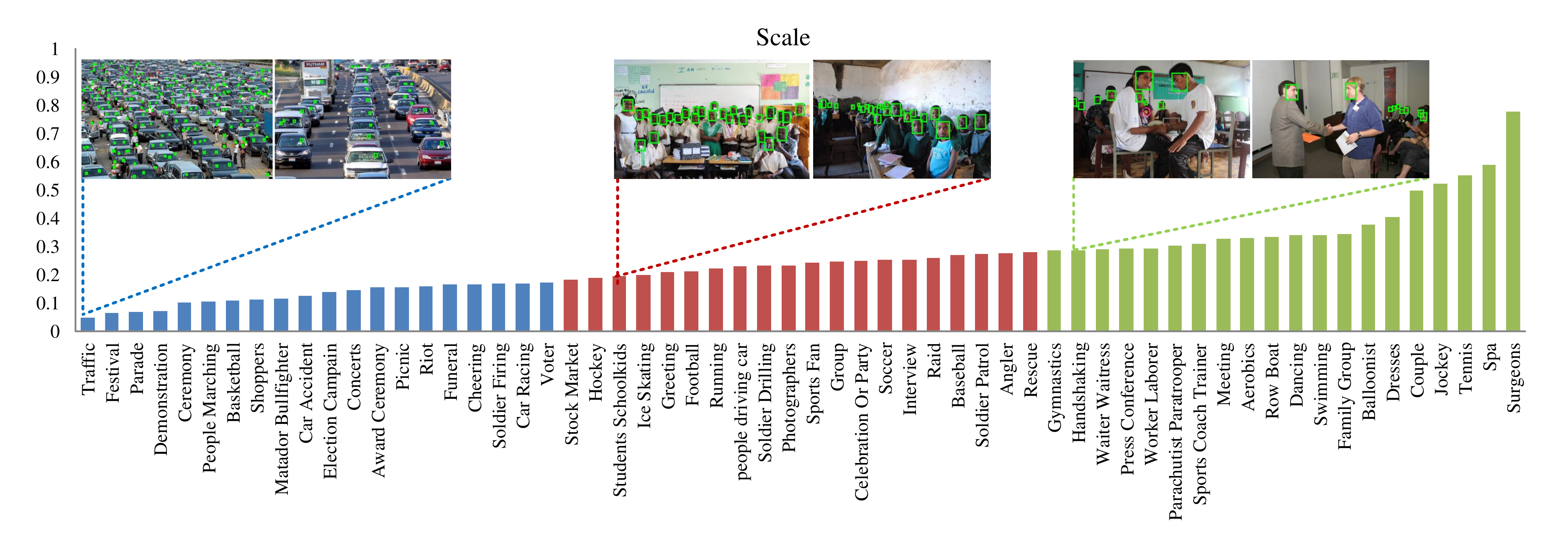}
\caption{\small{Histogram of detection rate for different event categories. Event categories are ranked in an ascending order based on the detection rate when the number of proposal is fixed at $10,000$. Top $1-20$, $21-40$, $41-60$ event categories are denoted in blue, red, and green, respectively. Example images for specific event classes are shown. Y-axis denotes for detection rate. X-axis denotes for event class name.}}
\label{fig:event_hist}
\end{center}
\end{figure*}

WIDER FACE dataset is challenging due to large variations in scale, occlusion, pose, and background clutters. These factors are essential to establishing the requirements for a real world system. To quantify these properties, we use generic object proposal approaches~\cite{mcg, uijlings2013selective, edgebox}, which are specially designed to discover potential objects in an image (face can be treated as an object). Through measuring the number of proposals vs. their detection rate of faces, we can have a preliminary assessment on the difficulty of a dataset and potential detection performance. In the following assessments, we adopt EdgeBox~\cite{edgebox} as object proposal, which has good performance in both accuracy and efficiency as evaluated in ~\cite{good_proposal}.

\noindent\textbf{Overall}. Fig.~\ref{fig:wider_stats}(a) shows that WIDER FACE has much lower detection rate compared with other face detection datasets. The results suggest that WIDER FACE is a more challenging face detection benchmark compared to existing datasets. Following the principles in KITTI~\cite{kitti} and MALF~\cite{yanfacedataset15} datasets, we define three levels of difficulty: 'Easy', 'Medium', 'Hard' based on the detection rate of EdgeBox~\cite{edgebox}, as shown in the Fig.~\ref{fig:wider_stats}(a). The average recall rates for these three levels are $92\%$, $76\%$, and $34\%$, respectively, with $8,000$ proposal per image.

%
\noindent\textbf{Scale}. We group the faces by their image size (height in pixels) into three scales: small (between $10$-$50$ pixels), medium (between $50$-$300$ pixels), large (over $300$ pixels). We make this division by considering the detection rate of generic object proposal and human performance.
%
%
As can be observed from Fig~\ref{fig:wider_stats}(b), the large and medium scales achieve high detection rate (more than $90\%$) with $8,000$ proposals per image. For the small scale, the detection rates consistently stay below $30\%$ even we increase the proposal number to $10,000$.

\noindent\textbf{Occlusion}. Occlusion is an important factor for evaluating the face detection performance. Similar to a recent study~\cite{yanfacedataset15}, we treat occlusion as an attribute and assign faces into three categories: no occlusion, partial occlusion, and heavy occlusion. Specifically, we ask annotator to measure the fraction of occlusion region for each face. A face is defined as `partially occluded' if $1\%$-$30\%$ of the total face area is occluded. A face with occluded area over $30\%$ is labeled as `heavily occluded'.
Fig.~\ref{fig:wider_anno} shows some examples of partial/heavy occlusions.
Fig.~\ref{fig:wider_stats}(c) shows that the detection rate decreases as occlusion level increases. The detection rates of faces with partial or heavy occlusions are below $50\%$ with $8,000$ proposals.

\noindent\textbf{Pose}. Similar to occlusion, we define two pose deformation levels, namely typical and atypical. Fig.~\ref{fig:wider_anno} shows some faces of typical and atypical pose. Face is annotated as atypical under two conditions: either the roll or pitch degree is larger than $30$-degree; or the yaw is larger than $90$-degree. Fig.~\ref{fig:wider_stats}(d) suggests that faces with atypical poses are much harder to be detected.

\begin{figure*}[t]
\begin{center}
\includegraphics[width=\linewidth]{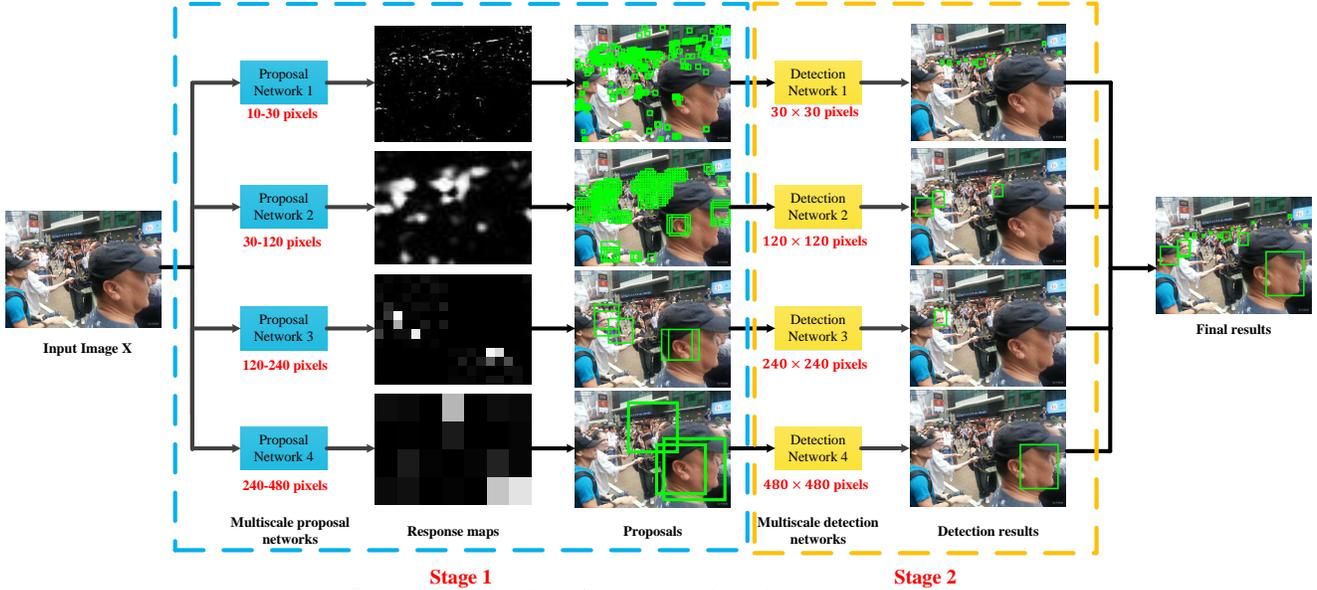}
\vskip -0.25cm
\caption{\small{The pipeline of the proposed multi-scale cascade CNN.}}
\vskip -0.25cm
\label{fig:multiscale_pipeline}
\end{center}
\end{figure*}   

\noindent\textbf{Event}. Different events are typically associated with different scenes. WIDER FACE contains $60$ event categories covering a large number of scenes in the real world, as shown in Fig.~\ref{fig:wider_intro} and Fig.~\ref{fig:wider_anno}. To evaluate the influence of event to face detection, we characterize each event with three factors: scale, occlusion, and pose. For each factor we compute the detection rate for the specific event class and then rank the detection rate in an ascending order. Based on the rank, events are divided into three partitions: easy ($41$-$60$ classes), medium ($21$-$40$ classes) and hard ($1$-$20$ classes). We show the partitions based on scale in Fig.~\ref{fig:event_hist}. Partitions based on occlusion and pose are included in the \textbf{appendix}.

\noindent\textbf{Effective training source}. As shown in the Table~\ref{tab:cmp_dataset}, existing datasets such as FDDB, AFW, and PASCAL FACE do not provide training data. Face detection algorithms tested on these datasets are frequently trained with ALFW~\cite{aflw}, which is designed for face landmark localization. However, there are two problems. First, ALFW omits annotations of many faces with a small scale, low resolution, and heavy occlusion. Second, the background in ALFW dataset is relatively clean. As a result, many face detection approaches resort to generate negative samples from other datasets such as PASCAL VOC dataset. 
In contrast, all recognizable faces are labeled in the WIDER FACE dataset. Because of its event-driven nature, WIDER FACE dataset has a large number of scenes with diverse background, making it possible as a good training source with both positive and negative samples. We demonstrate the effectiveness of WIDER FACE as a training source in Sec.~\ref{sec:training source}.

\section{Multi-scale Detection Cascade}
\label{sec:multi scale cascade}

We wish to establish a solid baseline for WIDER FACE dataset.
As we have shown in Table~\ref{tab:cmp_dataset}, WIDER FACE contains faces with a large range of scales. Fig.~\ref{fig:wider_stats}(b) further shows that faces with a height between $10$-$50$ pixels only achieve a proposal detection rate of below $30\%$. 
In order to deal with the high degree of variability in scale, we propose a multi-scale two-stage cascade framework and employ a divide and conquer strategy. Specifically, we train a set of face detectors, each of which only deals with faces in a relatively small range of scales. Each face detector consists of two stages. The first stage generates multi-scale proposals from a fully-convolutional network. The second stage is a multi-task convolutional network that generates face and non-face prediction of the candidate windows obtained from the first stage, and simultaneously predicts for face location. The pipeline is shown in Fig.~\ref{fig:multiscale_pipeline}. The two main steps are explained as follow.

\noindent\textbf{Multi-scale proposal}. In this step, we joint train a set of fully convolutional networks for face classification and scale classification. We first group faces into four categories by their image size, as shown in the Table~\ref{tab:scale_partition} (each row in the table represents a category). 
For each group, we further divide it into three subclasses. Each network is trained with image patches with the size of their upper bound scale. For example, Network $1$ and Network $2$ are trained with $30$$\times$$30$ and $120$$\times$$120$ image patches, respectively. 
We align a face at the center of an image patch as positive sample and assign a scale class label based on the predefined scale subclasses in each group. 
For negative samples, we randomly cropped patches from the training images. The patches should have an intersection-over-union (IoU) of smaller than $0.5$ with any of the positive samples. 
We assign a value $-1$ as the scale class for negative samples, which will have no contribution to the gradient during training. 

We take Network $2$ as an example. Let $\{\bx_i\}_{i=1}^N$ be a set of image patches where $\forall\bx_i\in\mathbb{R}^{120\times120}$. Similarly, let $\{\by^f_{i}\}_{i=1}^N$ be the set of face class labels and $\{\by^s_{i}\}_{i=1}^N$ be the set of scale class label, where $\forall\by^f_{i}\in\mathbb{R}^{1\times 1}$ and $\forall\by^s_{i}\in\mathbb{R}^{1\times 3}$. Learning is formulated as a multi-variate classification problem by minimizing the cross-entropy loss.
$L=\sum_{i=1}^{N}\by_i\log p(\mathbf{y}_i=1|\bx_i) + ({\mathbf{1}} - \by_i)\log\big({\mathbf{1}} - p(\by_i=1|\bx_i)\big)$,
where $p(\by_i|\bx_i)$ is modeled as a sigmoid function, indicating the probability of the presence of a face. This loss function can be optimized by the stochastic gradient descent with back-propagation.

\begin{table}[t]
\begin{center}
\caption{Summary of face scale for multi-scale proposal networks.}
\label{tab:scale_partition}
\vskip 0.15cm
\small\addtolength{\tabcolsep}{-1pt}
\addtolength{\tabcolsep}{-1pt}
\begin{tabular}{ c  || c | c | c}
\hline
\textbf{Scale} & Class 1 & Class 2 & Class 3 \\
\hline\hline
Network 1 & $10$-$15$ & $15$-$20$ & $20$-$30$\\
Network 2 & $30$-$50$ & $50$-$80$ & $80$-$120$\\
Network 3 & $120$-$160$ & $160$-$200$ & $200$-$240$\\
Network 4 & $240$-$320$ & $320$-$400$ & $400$-$480$\\
\hline
\end{tabular}
\end{center}
\vspace{-0.6cm}
\end{table}
\noindent\textbf{Face detection}. The prediction of proposed windows from the previous stage is refined in this stage.
For each scale category, we refine these proposals by joint training face classification and bounding box regression using the same CNN structure in the previous stage with the same input size. 
For face classification, a proposed window is assigned with a positive label if the IoU between it and the ground truth bounding box is larger than $0.5$; otherwise it is negative.
For bounding box regression, each proposal is predicted a position of its nearest ground truth bounding box.
If the proposed window is a false positive, the CNN outputs a vector of $[-1,-1,-1,-1]$. We adopt the Euclidean loss and cross-entropy loss for bounding box regression and face classification, respectively.
More details of face detection can be found in the \textbf{appendix}.

\section{Experimental Results}
\label{sec:experiments}
\subsection{Benchmarks}
\label{sec:benchmarks}

\begin{figure*}[t]
\begin{center}
{\includegraphics[width=\linewidth]{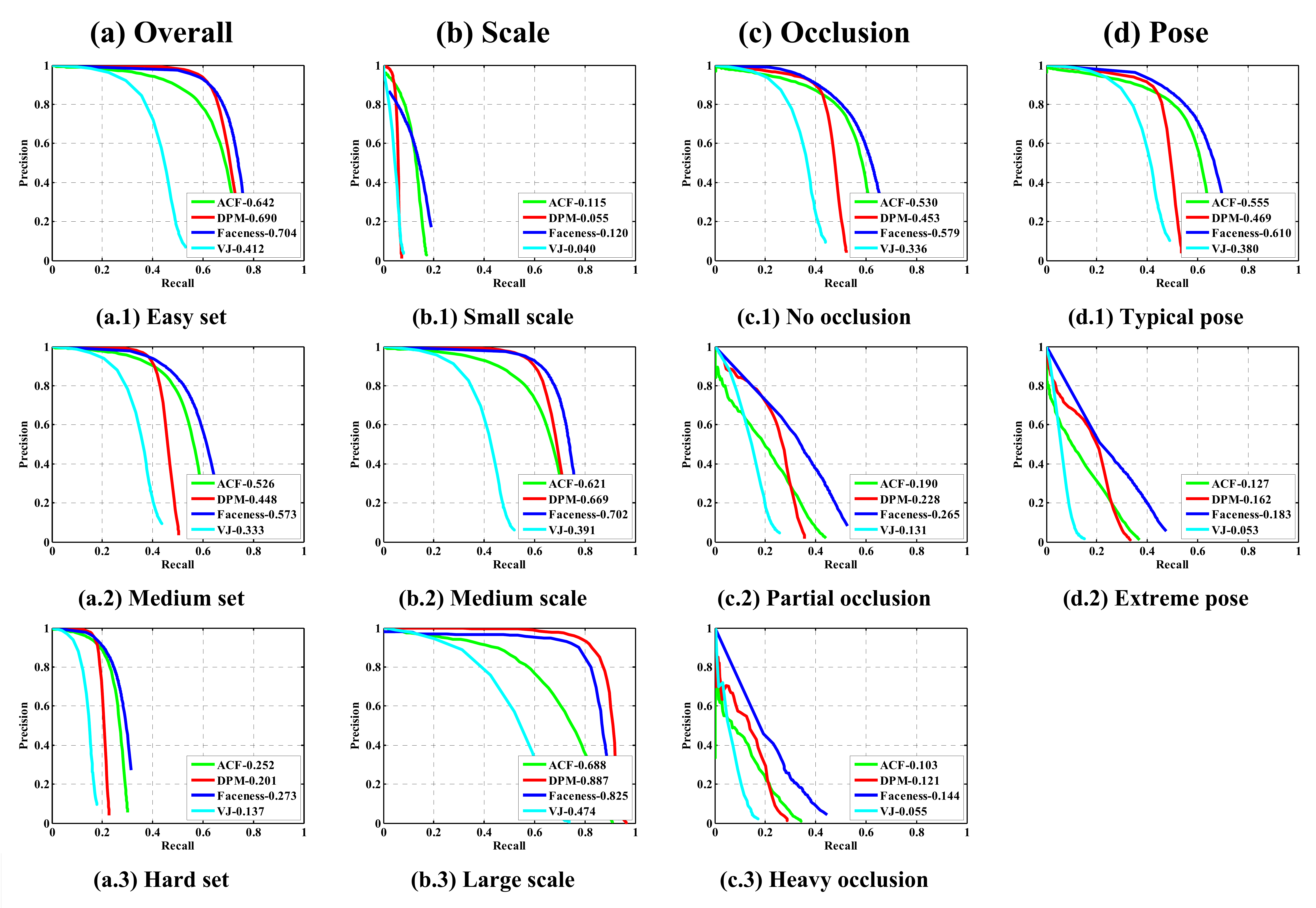}}
{\caption{\small{Precision and recall curves of different subsets of WIDER FACES: (a.1)-(a.3)  Overall Easy/Medium/Hard subsets. (b.1)-(b.3) Small/Medium/Large scale subsets. (c.1)-(c.3) None/Partial/Heavy occlusion subsets. (d.1)-(d.2) Typical/Atypical pose subsets.}}
\label{fig:wider_baselie}}
\end{center}
\end{figure*}

As we discussed in Sec.~\ref{sec:related work}, face detection algorithms can be broadly grouped into four representative categories.
For each class, we pick one algorithm as a baseline method. We select VJ~\cite{viola2004robust}, ACF~\cite{ACF-multiscale}, DPM~\cite{HeadHunter}, and Faceness~\cite{yang2015faceness} as baselines. The VJ~\cite{viola2004robust}, DPM~\cite{HeadHunter}, and Faceness~\cite{yang2015faceness} detectors are either obtained from the authors or from open source library (OpenCV). The ACF~\cite{ACF-multiscale} detector is reimplemented using the open source code. We adopt the Scenario-Ext here (see Sec.~\ref{sec:overview}), that is, these detectors were trained by using external datasets and are used `as is' without re-training them on WIDER FACE.
We employ PASCAL VOC~\cite{pascalvoc} evaluation metric for the evaluation. Following previous work~\cite{HeadHunter}, we conduct linear transformation for each method to fit the annotation of WIDER FACE.

\noindent\textbf{Overall}. In this experiment, we employ the evaluation setting mentioned in Sec.~\ref{sec:properties}. The results are shown in Fig.~\ref{fig:wider_baselie} (a.1)-(a.3). Faceness~\cite{yang2015faceness} outperforms other methods on three subsets, with DPM~\cite{HeadHunter} and ACF~\cite{ACF-multiscale} as marginal second and third. For the easy set, the average precision (AP) of most methods are over $60\%$, but none of them surpasses $75\%$.  The performance drops $10\%$ for all methods on the medium set. The hard set is even more challenging. The performance quickly decreases, with a AP below $30\%$ for all methods. To trace the reasons of failure, we examine performance on varying subsets of the data.\\
\noindent\textbf{Scale}. As described in Sec.~\ref{sec:properties}, we group faces according to the image height: small ($10$-$50$ pixels), medium ($50$-$300$ pixels), and large ($300$ or more pixels) scales. 
Fig.~\ref{fig:wider_baselie} (b.1)-(b.3) show the results for each scale on un-occluded faces only. For the large scale, DPM and Faceness obtain over $80\%$ AP. At the medium scale, Faceness achieves the best relative result but the absolute performance is only $70\%$ AP. The results of small scale are abysmal: none of the algorithms is able to achieve more than $12\%$ AP. This shows that current face detectors are incapable to deal with faces of small scale.\\  
\noindent\textbf{Occlusion}. Occlusion handling is a key performance metric for any face detectors. 
In Fig.~\ref{fig:wider_baselie} (c.1)-(c.3), we show the impact of occlusion on detecting faces with a height of at least $30$ pixels. As mentioned in Sec.~\ref{sec:properties}, we classify faces into three categories: un-occluded, partially occluded ($1\%$-$30\%$ area occluded) and heavily occluded (over $30\%$ area occluded). With partial occlusion, the performance drops significantly. The maximum AP is only $26.5\%$ achieved by Faceness. The performance further decreases in the heavy occlusion setting. The best performance of baseline methods drops to $14.4\%$. It is worth noting that Faceness and DPM, which are part based models, already perform relatively better than other methods on occlusion handling.\\
\noindent\textbf{Pose}. As discussed in Sec.~\ref{sec:properties}, we assign a face pose as atypical if either the roll or pitch degree is larger than $30$-degree; or the yaw is larger than $90$-degree. 
Otherwise a face pose is classified as typical. We show results in Fig.~\ref{fig:wider_baselie} (d.1)-(d.2). Faces which are un-occluded and with a scale larger than $30$ pixels are used in this experiment. The performance clearly degrades for atypical pose. The best performance is achieved by Faceness, with a recall below $20\%$. The results suggest that current face detectors are only capable of dealing with faces with out-of-plane rotation and a small range of in-plane rotation.

\begin{table*}[tp]
\centering
\caption{Comparison of per class AP. 
To save space, we only show abbreviations of category names here. The event category is organized based on the rank sequence in Fig.~\ref{fig:event_hist} (from hard to easy events based on scale measure). 
We compare the accuracy of Faceness and ACF models retrained on WIDER FACE training set with the baseline Faceness and ACF. 
With the help of WIDER FACE dataset, accuracies on $56$ out of $60$ categories have been improved. The re-trained Faceness model wins $30$ out of $60$ classes, followed by the ACF model with $26$ classes. Faceness wins $1$ medium class and $3$ easy classes.}  
\vspace{0.2cm}
\footnotesize                                                                                    \setlength{\tabcolsep}{1.9pt}                                                       
\begin{tabulary}{\linewidth}{lcccccccccccccccccccc}                                                                                                                                                   \toprule[1pt]                                                                                                                             
& Traf. & Fest. & Para. & Demo. & Cere. & March. & Bask. & Shop. & Mata. & Acci. & Elec. & Conc. & Awar. & Picn. & Riot. & Fune. & Chee. & Firi. & Raci. & Vote. \\     \midrule                                                                                                                                                              
ACF &  $.421$ &  $.368$ &  $.431$ &  $.330$ &  $.521$ &  $.381$ &  $.452$ &  $.503$ &  $.308$ &  $.254$ &  $.409$ &  $.512$ &  $.720$ &  $.475$ &  $.388$ &  $.502$ &  $.474$ &  $.320$ &  $.552$ &  $.457$ \\
ACF-WIDER &  $.385$ &  $.435$ &  $.528$ &  $\mathbf{.464}$ &  $.595$ &  $.490$ &  $\mathbf{.562}$ &  $\mathbf{.603}$ &  $.334$ &  $.352$ &  $\mathbf{.538}$ &  $.486$ &  $\mathbf{.797}$ &  $.550$ &  $.395$ &  $.568$ &  $\mathbf{.589}$ &  $.432$ &  $\mathbf{.669}$ &  $.532$ \\
Faceness &  $.497$ &  $.376$ &  $.459$ &  $.410$ &  $.547$ &  $.434$ &  $.481$ &  $.575$ &  $.388$ &  $.323$ &  $.461$ &  $.569$ &  $.730$ &  $.526$ &  $.455$ &  $.563$ &  $.496$ &  $.439$ &  $.577$ &  $.535$ \\
Faceness-WIDER &  $\mathbf{.535}$ &  $\mathbf{.451}$ &  $\mathbf{.560}$ &  $.454$ &  $\mathbf{.626}$ &  $\mathbf{.495}$ &  $.525$ &  $.593$ &  $\mathbf{.432}$ &  $\mathbf{.358}$ &  $.489$ &  $\mathbf{.576}$ &  $.737$ &  $\mathbf{.621}$ &  $\mathbf{.486}$ &  $\mathbf{.579}$ &  $.555$ &  $\mathbf{.454}$ &  $.635$ &  $\mathbf{.558}$ \\

\bottomrule
\toprule                                                                                                                               
& Stoc. & Hock. & Stud. & Skat. & Gree. & Foot. & Runn. & Driv. & Dril. & Phot. & Spor. & Grou. & Cele. & Socc. & Inte. & Raid. & Base. & Patr. & Angl. & Resc.\\
\midrule                                                                                                                                                          
ACF &  $.549$ &  $.430$ &  $.557$ &  $.502$ &  $.467$ &  $.394$ &  $.626$ &  $.562$ &  $.447$ &  $.576$ &  $.343$ &  $.685$ &  $.577$ &  $.719$ &  $.628$ &  $.407$ &  $.442$ &  $.497$ &  $.564$ &  $.465$ \\
ACF-WIDER  &  $.519$ &  $\mathbf{.591}$ &  $\mathbf{.666}$ &  $\mathbf{.630}$ &  $.546$ &  $\mathbf{.508}$ &  $\mathbf{.707}$ &  $.609$ &  $\mathbf{.521}$ &  $.627$ &  $\mathbf{.430}$ &  $.756$ &  $.611$ &  $.727$ &  $.616$ &  $\mathbf{.506}$ &  $.583$ &  $.529$ &  $\mathbf{.645}$ &  $.546$ \\
Faceness &  $\mathbf{.617}$ &  $.481$ &  $.639$ &  $.561$ &  $.576$ &  $.475$ &  $.667$ &  $.643$ &  $.469$ &  $.628$ &  $.406$ &  $.725$ &  $.563$ &  $.744$ &  $.680$ &  $.457$ &  $.499$ &  $.538$ &  $.621$ &  $.520$ \\
Faceness-WIDER &  $.611$ &  $.579$ &  $.660$ &  $.599$ &  $\mathbf{.588}$ &  $.505$ &  $.672$ &  $\mathbf{.648}$ &  $.519$ &  $\mathbf{.650}$ &  $.409$ &  $\mathbf{.776}$ &  $\mathbf{.621}$ &  $\mathbf{.768}$ &  $\mathbf{.686}$ &  $.489$ &  $\mathbf{.607}$ &  $\mathbf{.607}$ &  $.629$ &  $\mathbf{.564}$ \\
\bottomrule
\toprule                                                                                                         
& Gymn. & Hand. & Wait. & Pres. & Work. & Parach. & Coac. & Meet. & Aero. & Boat. & Danc. & Swim. & Fami. & Ball. & Dres. & Coup. & Jock. & Tenn. & Spa. & Surg. \\       \midrule                                                                                                                        
ACF &  $.749$ &  $.472$ &  $.722$ &  $.720$ &  $.589$ &  $.435$ &  $.598$ &  $.548$ &  $.629$ &  $.530$ &  $.507$ &  $.626$ &  $.755$ &  $.589$ &  $.734$ &  $.621$ &  $.667$ &  $.701$ &  $.386$ &  $.599$ \\
ACF-WIDER  &  $.750$ &  $\mathbf{.589}$ &  $\mathbf{.836}$ &  $\mathbf{.794}$ &  $\mathbf{.649}$ &  $.492$ &  $\mathbf{.705}$ &  $\mathbf{.700}$ &  $\mathbf{.734}$ &  $.602$ &  $.524$ &  $.534$ &  $\mathbf{.856}$ &  $.642$ &  $\mathbf{.802}$ &  $.589$ &  $\mathbf{.827}$ &  $.667$ &  $.418$ &  $.586$  \\
Faceness &  $.756$ &  $.540$ &  $.782$ &  $.732$ &  $.645$ &  $.517$ &  $.618$ &  $.592$ &  $.678$ &  $.569$ &  $.558$ &  $\mathbf{.666}$ &  $.809$ &  $.647$ &  $.774$ &  $\mathbf{.742}$ &  $.662$ &  $.744$ &  $.470$ &  $\mathbf{.635}$ \\
Faceness-WIDER &  $\mathbf{.768}$ &  $.577$ &  $.740$ &  $.746$ &  $.640$ &  $\mathbf{.540}$ &  $.637$ &  $.670$ &  $.718$ &  $\mathbf{.628}$ &  $\mathbf{.595}$ &  $.659$ &  $.842$ &  $\mathbf{.682}$ &  $.754$ &  $.699$ &  $.688$ &  $\mathbf{.759}$ &  $\mathbf{.493}$ &  $.632$ \\
\bottomrule[1pt]

\end{tabulary}                                                                                                                              
\label{table:class_acc}                                                                                                                                                                                      \vspace{-8pt}                               
\end{table*} 

\noindent\textbf{Summary}. Among the four baseline methods, Faceness tends to outperform the other methods. VJ performs poorly on all settings. DPM gains good performance on medium/large scale and occlusion. ACF outperforms DPM on small scale, no occlusion and typical pose settings. However, the overall performance is poor on WIDER FACE, suggesting a large room of improvement.

\subsection{WIDER FACE as an Effective Training Source}
\label{sec:training source}
In this experiment, we demonstrate the effectiveness of WIDER FACE dataset as a training source. We adopt Scenario-Int here (see Sec.~\ref{sec:overview}). We train ACF and Faceness on WIDER FACE to conduct this experiment. These two algorithms have shown relatively good performance on WIDER FACE previous benchmarks see (Sec.~\ref{sec:benchmarks}). 
Faces with a scale larger than $30$ pixels in the training set are used to retrain both methods. 
We train the ACF detector using the same training parameters as the baseline ACF. The negative samples are generated from the training images. For the Faceness detector, we first employ models shared by the authors to generate face proposals from the WIDER FACE training set. After that, we train the classifier with the same procedure described in ~\cite{yang2015faceness}. We test these models (denoted as ACF-WIDER and Faceness-WIDER) on WIDER FACE testing set and FDDB dataset.

\begin{figure}[t]
\begin{center}
\vskip -0.5cm
\vspace{-0.25cm}
\subfigure[WIDER Easy]{
	 \includegraphics[width=0.45\linewidth]{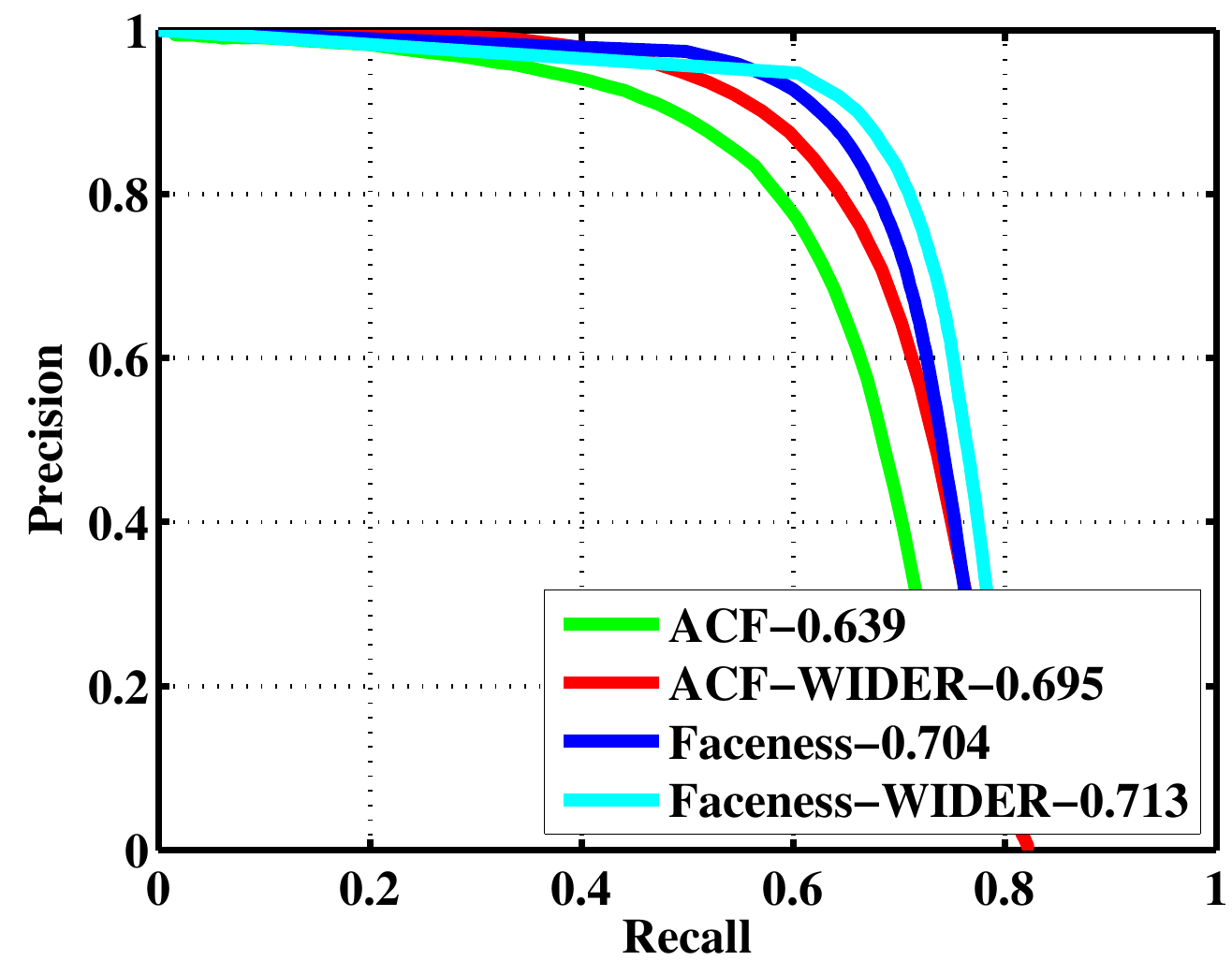}
}
\subfigure[WIDER Medium]{
	 \includegraphics[width=0.45\linewidth]{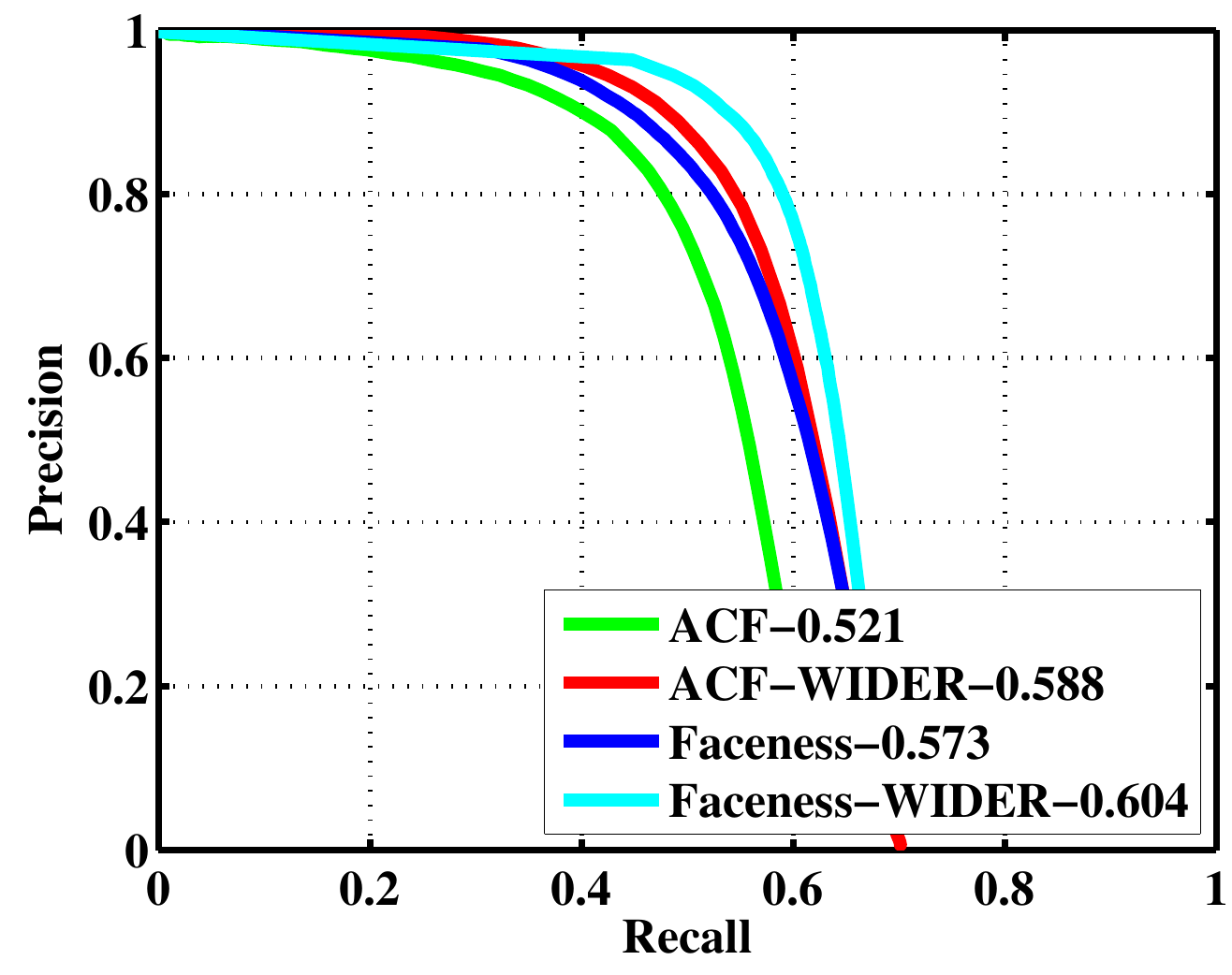}
}
\subfigure[WIDER Hard]{
	 \includegraphics[width=0.45\linewidth]{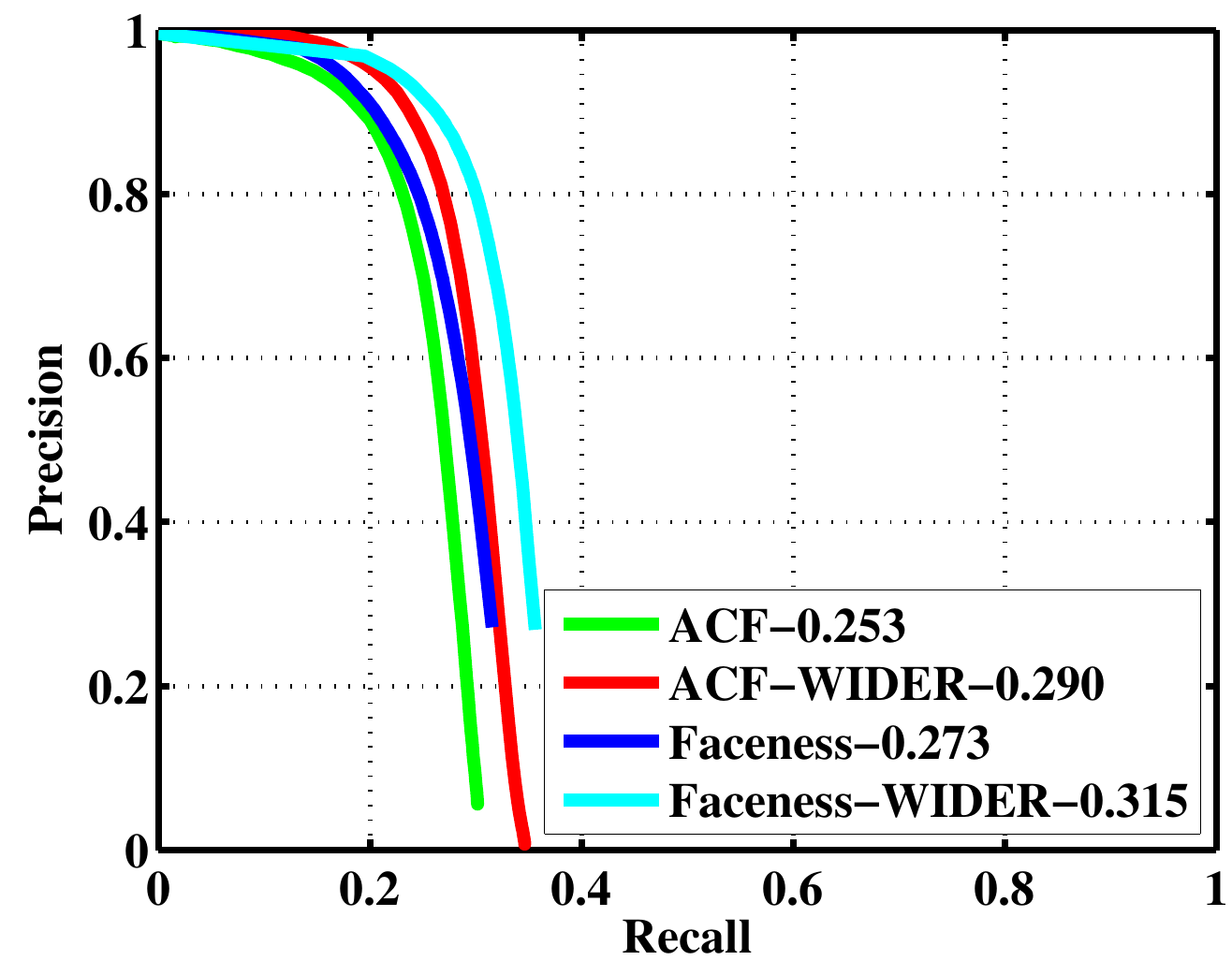}
}
\subfigure[FDDB]{
	 \includegraphics[width=0.47\linewidth]{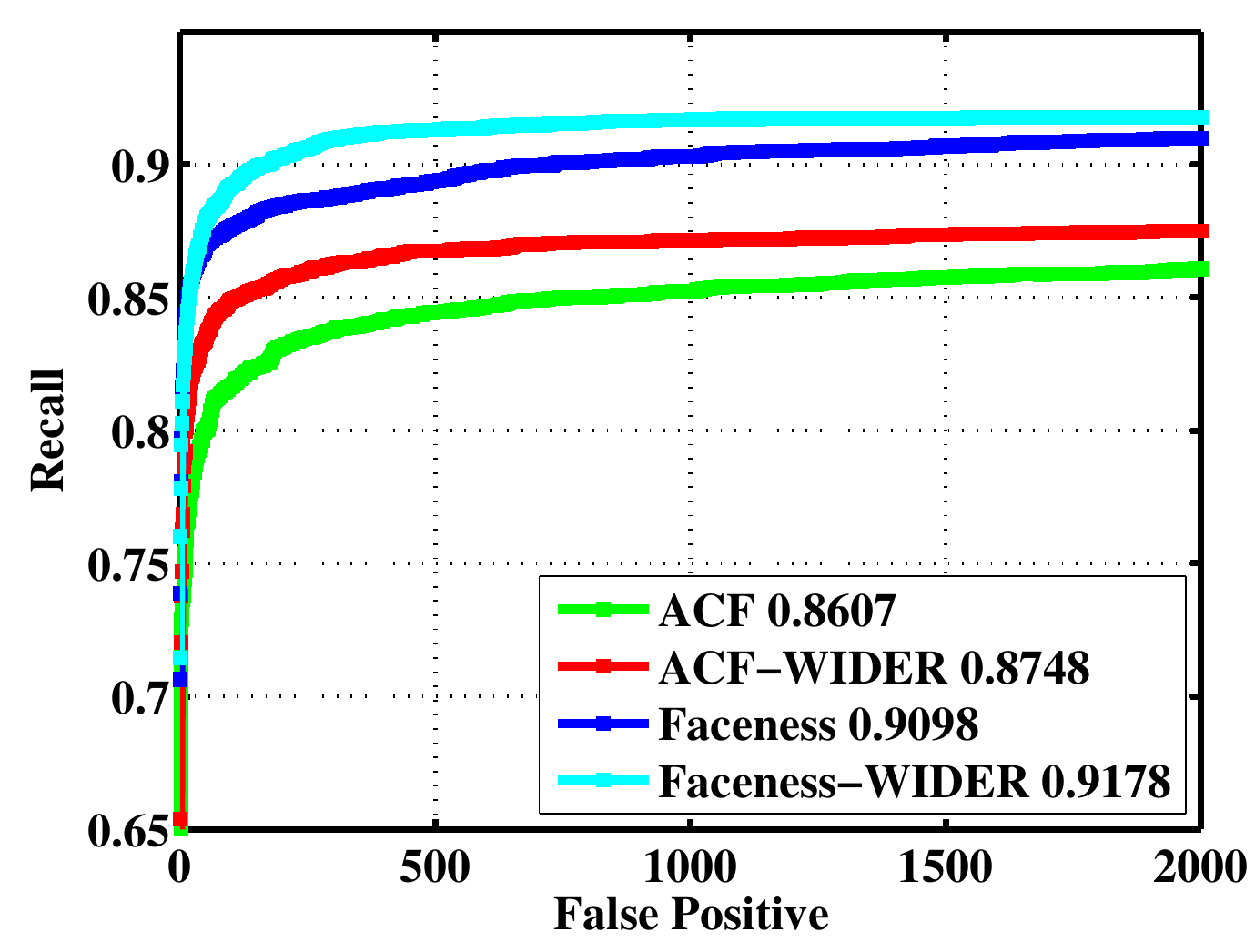}
}
\caption{\small{WIDER FACE as an effective training source. ACF-WIDER and Faceness-WIDER are retrained with WIDER FACE, while ACF and Faceness are the original models. (a)-(c) Precision and recall curves on WIDER Easy/Medium/Hard subsets. (d) ROC curve on FDDB dataset.}}
\vspace{-0.25cm}
\vskip -0.5cm
\label{fig:wider_retrain}
\end{center}
\end{figure}

\noindent\textbf{WIDER FACE}. As shown in Fig.~\ref{fig:wider_retrain}, the retrained models perform consistently better than the baseline models. The average AP improvement of retrained ACF detector is $5.4\%$ in comparison to baseline ACF detector. For the Faceness, the retrained Faceness model obtain $4.2\%$ improvement on WIDER hard test set.

\noindent\textbf{FDDB}. We further evaluate the retrained models on FDDB dataset. Similar to WIDER FACE dataset, the retrained models achieve improvement in comparison to the baseline methods. The retrained ACF detector achieves a recall rate of $87.48\%$, outperforms the baseline ACF by a considerable margin of $1.4\%$. The retrained Faceness detector obtains a high recall rate of $91.78\%$.  The recall rate improvement of the retrained Faceness detector is $0.8\%$ in comparison to the baseline Faceness detector. It worth noting that the retrained Faceness detector performs much better than the baseline Faceness detector when the number of false positive is less than $300$.\\
\noindent\textbf{Event}. 
%
We evaluate the baseline methods on each event class individually and report the results in Table~\ref{table:class_acc}. Faces with a height larger than $30$ pixels are used in this experiment. We compare the accuracy of Faceness and ACF models retrained on WIDER FACE training set with the baseline Faceness and ACF. With the help of WIDER FACE dataset, accuracies on $56$ out of $60$ event categories have been improved.
It is interesting to observe that the accuracy obtained highly correlates with the difficulty levels specified in Sec.~\ref{sec:properties} (also refer to Fig.~\ref{fig:event_hist}). For example, the best performance on "Festival" which is assigned as a hard class is no more than $46\%$ AP. 

\subsection{Evaluation of Multi-scale Detection Cascade}
\label{sec:multiscale cascaed exp}
In this experiment we evaluate the effectiveness of the proposed multi-scale cascade algorithm. 
Apart from the ACF-WIDER and Faceness-WIDER models (Sec.~\ref{sec:training source}), we establish a baseline based on a "Two-stage CNN". This model differs to our multi-scale cascade model in the way it handles multiple face scales. Instead of having multiple networks targeted for different scales, the two-stage CNN adopts a more typical approach. Specifically, its first stage consists only a single network to perform face classification. During testing, an image pyramid that encompasses different scales of a test image is fed to the first stage to generate multi-scale face proposals. The second stage is similar to our multi-scale cascade model -- it performs further refinement on proposals by simultaneous face classification and bounding box regression.

\begin{figure}[t]
\begin{center}
\vskip -0.5cm
\vspace{-0.25cm}
\subfigure[WIDER Easy]{
	 \includegraphics[width=0.45\linewidth]{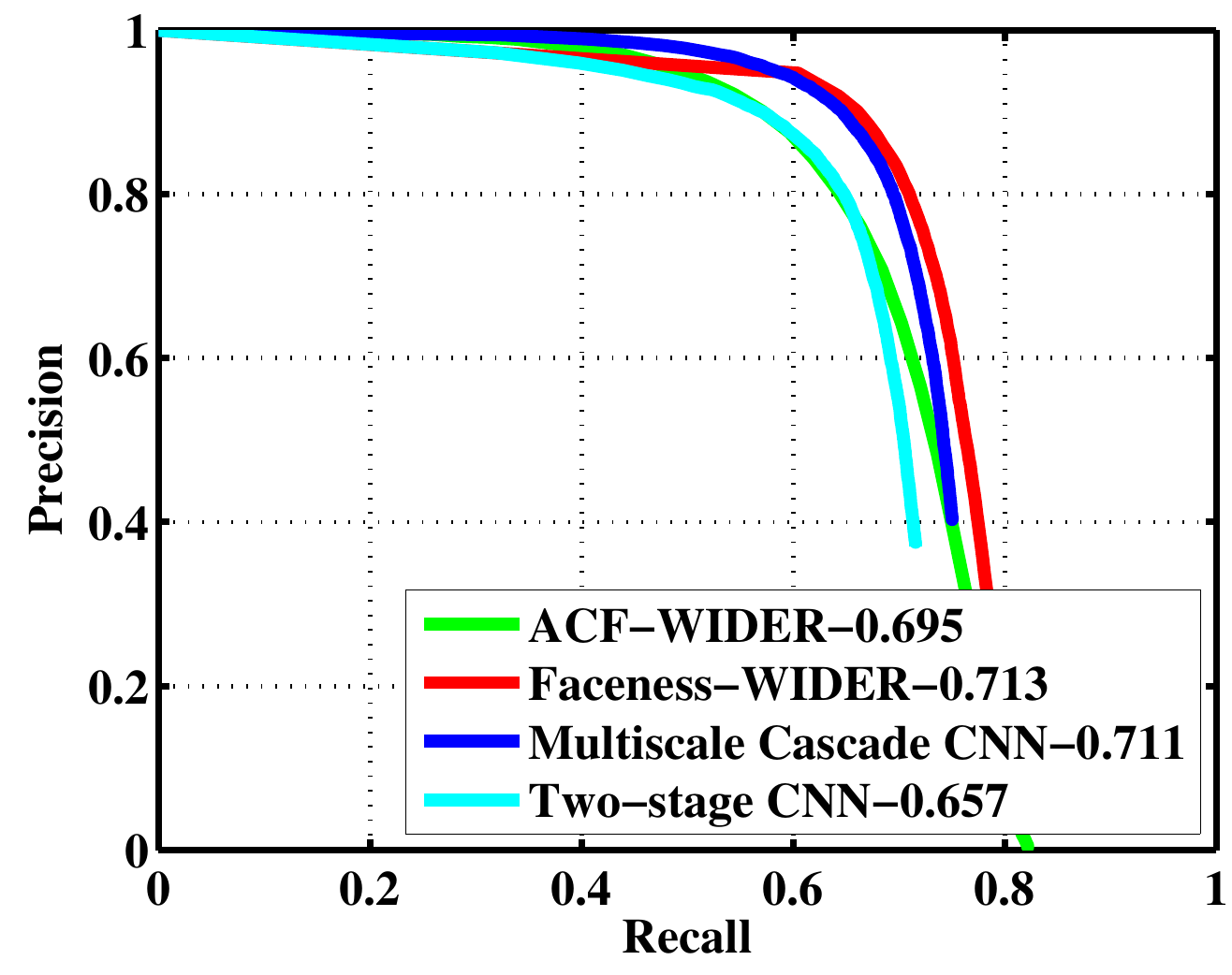}
}
\subfigure[WIDER Medium]{
	 \includegraphics[width=0.45\linewidth]{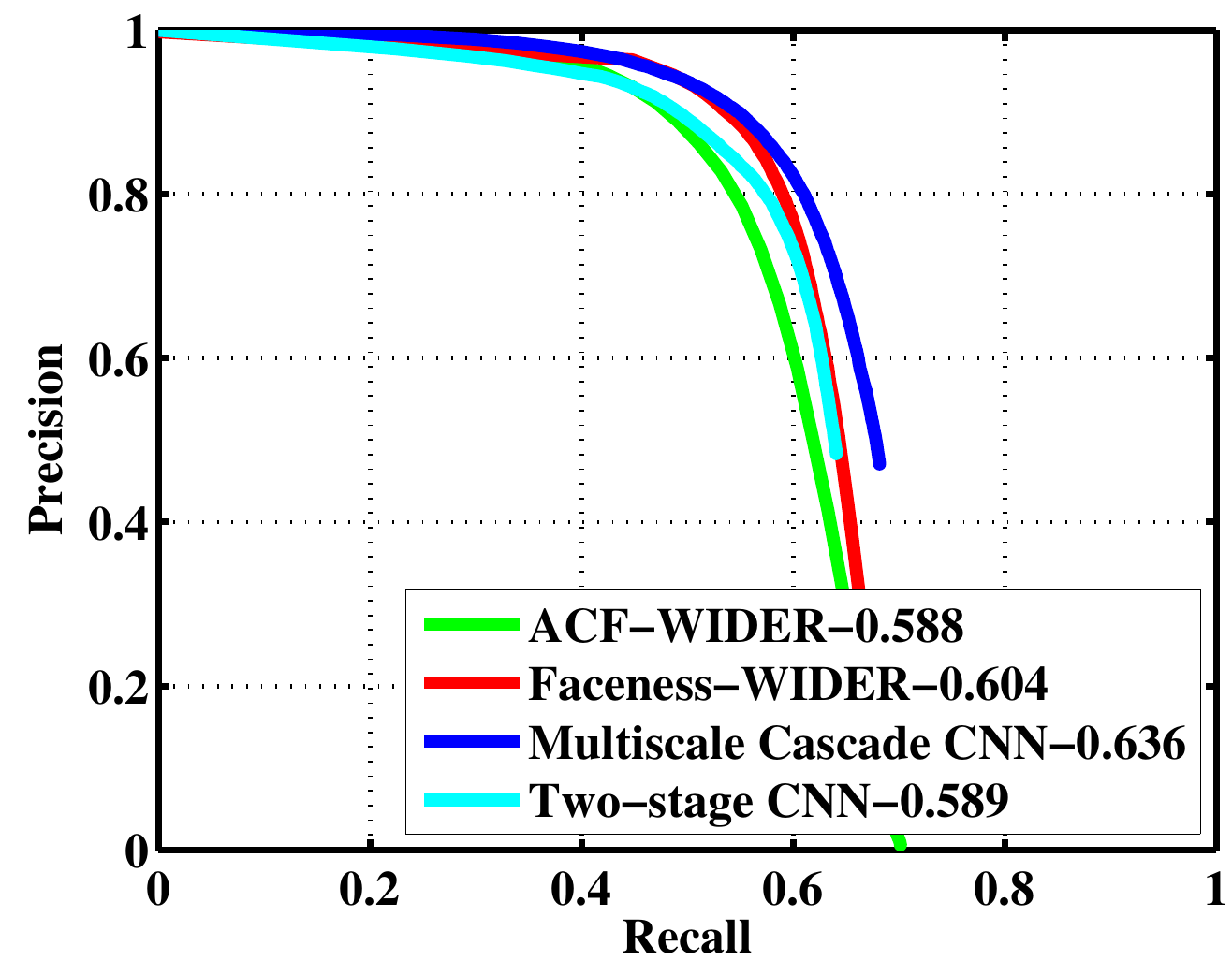}
}
\subfigure[WIDER Hard]{
	 \includegraphics[width=0.45\linewidth]{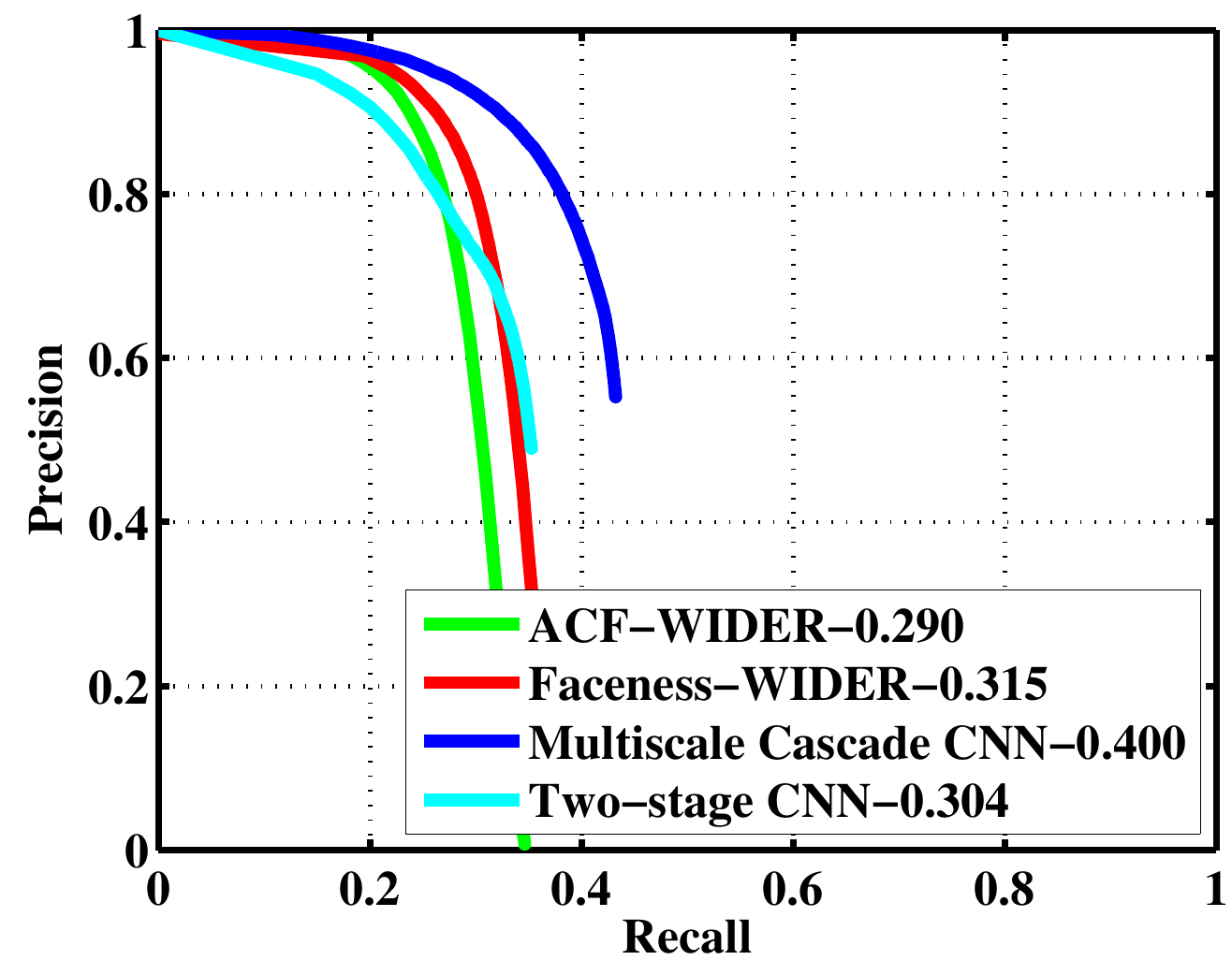}
}
\caption{\small{Evaluation of multi-scale detection cascade: (a)-(c) Precision and recall curves on WIDER Easy/Medium/Hard subsets.}}
\vspace{-0.25cm}
\vskip -0.5cm
\label{fig:multi-scale cascade}
\end{center}
\end{figure}

We evaluate the multi-scale cascade CNN and baseline methods on WIDER Easy/Medium/Hard subsets. As shown in Fig.~\ref{fig:multi-scale cascade}, the multi-scale cascade CNN obtains $8.5\%$ AP improvement on the WIDER Hard subset compared to the retrained Faceness, suggesting its superior capability in handling faces with different scales. In particular, having multiple networks specialized on different scale range is shown effective in comparison to using a single network to handle multiple scales. In other words, it is difficult for a single network to handle large appearance variations caused by scale. For the WIDER Medium subset, the multi-scale cascade CNN outperforms other baseline methods with a considerable margin. All models perform comparably on the WIDER Easy subset.

\section{Conclusion}
\label{sec:conclusion}

We have proposed a large, richly annotated WIDER FACE dataset for training and evaluating face detection algorithms. We benchmark four representative face detection methods. Even considering an easy subset (typically with faces of over $50$ pixels height), existing state-of-the-art algorithms reach only around $70\%$ AP, as shown in Fig.~\ref{fig:multi-scale cascade}.
With this new dataset, we wish to encourage the community to focusing on some inherent challenges of face detection -- small scale, occlusion, and extreme poses. These factors are ubiquitous in many real world applications. For instance, faces captured by surveillance cameras in public spaces or events are typically small, occluded, and atypical poses. These faces are arguably the most interesting yet crucial to detect for further investigation.
%

%

\newpage
\section{Appendix}

\subsection{Multi-scale Detection Cascade}
\label{sec:multi scale cascade}
Multi-scale detection cascade CNN consists of a set of face detectors, with each of them only deals with faces in a relatively small range of scale. Each face detector consists of two stages. The first stage generates multi-scale proposal from a fully-convolutional network. The second stage gives face and non-face prediction of the candidate windows generate from first stage. If the candidate window is classified as face, we further refine the location of the candidate window.
\vspace{-0.3cm}
\subsubsection{Training Multi-scale Proposal Network}
We provide details of the training process for multi-scale proposal networks. In this step, we train four fully convolutional networks for face classification and scale classification. The network structures are summarized in Table~\ref{tab:Network 1}, Table~\ref{tab:Network 2}, Table~\ref{tab:Network 3}, and Table~\ref{tab:Network 4}. As we described in the paper, we group faces into four categories by their image size, as shown in the Table~\ref{tab:scale_partition} (each row in the table represents a category). For each group, we further divide it into three subclasses. Each network is trained with image patches with the size of their upper bound scale. For example, Proposal Network $1$ and Proposal Network $2$ are trained with $30$$\times$$30$ and $120$$\times$$120$ image patches respectively.
For the Proposal Network $1$, Proposal Network $2$, and Proposal Network $3$, we initialize the layers from Conv $1$ to Conv $5$ using the Imagenet $1,000$ categories pre-trained Clarifai net. For the Proposal Network $4$, we initialize the layers from Conv $2$ to Conv $5$ using pre-trained Clarifai net. The remaining layers in each network are randomly initialized with weights drawn from a Gaussian distribution of $\mu = 0$ and $\sigma = 0.01$.
To account for the multi-label scenario, cross-entropy loss is adopted as shown below:
\begin{equation}\label{eq:finetune}
\begin{split}
&L=\sum_{i=1}^{|D|}(\mathbf{y_i}\log p(\mathbf{y_i}|\mathbf{I_i}) + ({\mathbf{1}} - \mathbf{y_i})\log({\mathbf{1}} - p(\mathbf{y_i}|\mathbf{I_i}))),~~~\mathrm{and}\\
&p(\mathbf{y_i}=c|\mathbf{I_i})=\frac{{\mathbf{1}}}{{\mathbf{1}} + \exp(-f(\mathbf{I_i}))},
\end{split}
\end{equation}
Back propagation and SGD are also employed here for optimizing Eqn.~(\ref{eq:finetune}). Similar to \cite{RCNN, imagenet}, we set the initial fine-tuning learning rate as one-tenth of the corresponding pre-training learning rate and drop it by a factor of $10$ throughout training. After training, we conduct hard negative mining on the training set and further tune the proposal networks using hard negative samples. 

\begin{table}[t]
\begin{center}
\caption{Summary of face scale for multi-scale proposal networks.}
\label{tab:scale_partition}
\vskip 0.15cm
\small\addtolength{\tabcolsep}{-1pt}
\addtolength{\tabcolsep}{-1pt}
\begin{tabular}{ c  || c | c | c}
\hline
\textbf{Scale} & Class 1 & Class 2 & Class 3 \\
\hline\hline
Network 1 & $10$-$15$ & $15$-$20$ & $20$-$30$\\
Network 2 & $30$-$50$ & $50$-$80$ & $80$-$120$\\
Network 3 & $120$-$160$ & $160$-$200$ & $200$-$240$\\
Network 4 & $240$-$320$ & $320$-$400$ & $400$-$480$\\
\hline
\end{tabular}
\end{center}
\vspace{-0.6cm}
\end{table}

\begin{table}[t]
\begin{center}
\caption{Model structure of Network 1.}
\label{tab:Network 1}
\vskip 0.15cm
\tiny\addtolength{\tabcolsep}{-1pt}
\addtolength{\tabcolsep}{-1pt}
\begin{tabular}{ c  || c | c | c | c| c}
\hline
\textbf{Layer name} & Filter Number & Filter Size & Stride & Padding & Activation Function \\
\hline\hline
Conv 1 & 96 & 7$\times$7 & 2 & 0 & RELU\\
Pool 1 & - & 3$\times$3 & 2 & - & -\\
LRN 1 & - & 5$\times$5 & - & - & -\\
Conv 2 & 256 & 5$\times$5 & 1 & 1 & RELU\\
LRN 2 & - & 5$\times$5 & - & - & -\\
Conv 3 & 384 & 3$\times$3 & 1 & 1 & RELU\\
Conv 4 & 384 & 3$\times$3 & 1 & 1 & RELU\\
Conv 5 & 256 & 3$\times$3 & 1 & 1 & RELU\\
Conv 6 & 256 & 4$\times$4 & 1 & 0 & RELU\\
Conv 7 & 256 & 1$\times$1 & 1 & 0 & RELU\\
\hline
\end{tabular}
\end{center}
\vspace{-0.6cm}
\end{table}

\begin{table}[t]
\begin{center}
\caption{Model structure of Network 2.}
\label{tab:Network 2}
\vskip 0.15cm
\tiny\addtolength{\tabcolsep}{-1pt}
\addtolength{\tabcolsep}{-1pt}
\begin{tabular}{ c  || c | c | c | c | c }
\hline
\textbf{Layer name} & Filter Number & Filter Size & Stride & Padding & Activation Function \\
\hline\hline
Conv 1 & 96 & 7$\times$7 & 2 & 0& RELU\\
Pool 1 & - & 3$\times$3 & 2 & - & -\\
LRN 1 & - & 5$\times$5 & - & - & -\\
Conv 2 & 256 & 5$\times$5 & 2 & 1 & RELU\\
Pool 2 & - & 3$\times$3 & 2 & - & -\\
LRN 2 & - & 5$\times$5 & - & - & -\\
Conv 3 & 384 & 3$\times$3 & 1 & 1& RELU\\
Conv 4 & 384 & 3$\times$3 & 1 & 1& RELU\\
Conv 5 & 256 & 3$\times$3 & 1 & 1& RELU\\
Conv 6 & 256 & 3$\times$3 & 1 & 1& RELU\\
Conv 7 & 256 & 3$\times$3 & 1 & 0& RELU\\
Conv 8 & 1024 & 4$\times$4 & 1 & 0& RELU\\
Conv 9 & 1024 & 1$\times$1 & 1 & 0& RELU\\
\hline
\end{tabular}
\end{center}
\vspace{-0.6cm}
\end{table}

\begin{table}[t]
\begin{center}
\caption{Model structure of Network 3.}
\label{tab:Network 3}
\vskip 0.15cm
\tiny\addtolength{\tabcolsep}{-1pt}
\addtolength{\tabcolsep}{-1pt}
\begin{tabular}{ c  || c | c | c | c | c }
\hline
\textbf{Layer name} & Filter Number & Filter Size & Stride & Padding & Activation Function \\
\hline\hline
Conv 1 & 96 & 7$\times$7 & 2 & 0& RELU\\
Pool 1 & - & 3$\times$3 & 2 & - & -\\
LRN 1 & - & 5$\times$5 & - & - & -\\
Conv 2 & 256 & 5$\times$5 & 2 & 1 & RELU\\
Pool 2 & - & 3$\times$3 & 2 & - & -\\
LRN 2 & - & 5$\times$5 & - & - & -\\
Conv 3 & 384 & 3$\times$3 & 1 & 1& RELU\\
Conv 4 & 384 & 3$\times$3 & 1 & 1& RELU\\
Conv 5 & 256 & 3$\times$3 & 1 & 1& RELU\\
Pool 5 & - & 3$\times$3 & 3 & - & -\\
Conv 6 & 4096 & 5$\times$5 & 1 & 0& RELU\\
Conv 7 & 4096 & 1$\times$1 & 1 & 0& RELU\\
\hline
\end{tabular}
\end{center}
\vspace{-0.6cm}
\end{table}

\begin{table}[t]
\begin{center}
\caption{Model structure of Network 4.}
\label{tab:Network 4}
\vskip 0.15cm
\tiny\addtolength{\tabcolsep}{-1pt}
\addtolength{\tabcolsep}{-1pt}
\begin{tabular}{ c  || c | c | c | c | c }
\hline
\textbf{Layer name} & Filter Number & Filter Size & Stride & Padding & Activation Function \\
\hline\hline
Conv 1 & 96 & 11$\times$11 & 4 & 0& RELU\\
Pool 1 & - & 3$\times$3 & 2 & - & -\\
LRN 1 & - & 5$\times$5 & - & - & -\\
Conv 2 & 256 & 5$\times$5 & 2 & 1 & RELU\\
Pool 2 & - & 3$\times$3 & 2 & - & -\\
LRN 2 & - & 5$\times$5 & - & - & -\\
Conv 3 & 384 & 3$\times$3 & 1 & 1& RELU\\
Conv 4 & 384 & 3$\times$3 & 1 & 1& RELU\\
Conv 5 & 256 & 3$\times$3 & 1 & 1& RELU\\
Pool 5 & - & 3$\times$3 & 2 & - & -\\
Conv 6 & 4096 & 4$\times$4 & 1 & 0& RELU\\
Conv 7 & 4096 & 1$\times$1 & 1 & 0& RELU\\
\hline
\end{tabular}
\end{center}
\vspace{-0.6cm}
\end{table}

\subsubsection{Training Face Detector}
In this section, we provide more details of the training process for face detection. As mentioned at beginning, we train four fully convolutional neural networks for face detection. Each detection network is trained to conduct face classification and bounding box regression simultaneously. We fine-tune the detection networks in this stage using respective proposal network in the previous stage. For example, the Detection Network 1 is trained by fine-tuning the Proposal Network 1 in the previous stage with the same structure. The detection networks are trained with face proposals generated from respective proposal networks. For face classification, we assign a proposed bounding box to a ground truth bounding box based on the minimal Euclidean distances between the center of the proposed bounding box and the center of ground truth bounding box. 
The proposed bounding box is assigned as positive, if the IoU between proposed bounding box and the assigned ground truth bounding box is larger than $0.5$; otherwise it is negative.
For bounding box regression, we train the multi-task deep convolutional neural network to regress each proposal to predict the positions of its assigned ground truth bounding box.

During the training process, if the number of positive samples is less than $10\%$ of the total samples, we randomly cropped the ground truth faces and add these samples as positive samples. 
We adopt Euclidean loss and cross-entropy loss for bounding box regression and face classification, respectively.

\subsection{WIDER FACE Dataset}
\label{sec:dataset}
\noindent\textbf{Event}. We measure each event with three factors: scale, occlusion and pose. For each factor, we compute the detection rate for the specific event class and then rank the detection rate in the ascending order. Three classes are divided: easy (top $41$-$60$ class), medium (top $21$-$40$ class) and hard (top $1$-$20$ class), as shown in the Fig.~\ref{fig:event_hist}. The corresponding relationship between abbreviations of the events categories and full name of the event categories are shown in Table~\ref{table:class_name}. 

\begin{figure*}[t]
\begin{center}
\includegraphics[width=\linewidth]{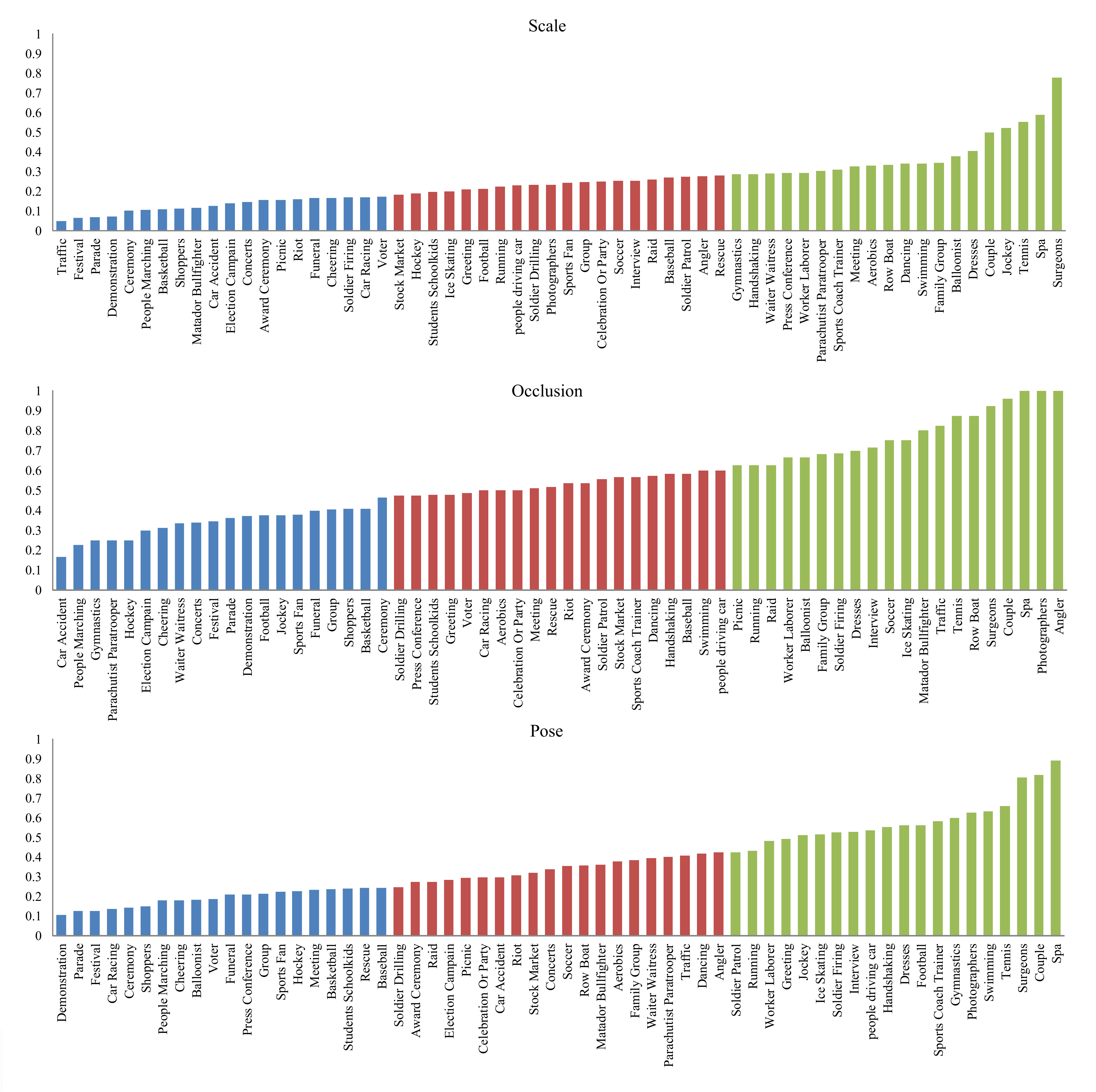}
\vskip -0.25cm
\caption{\small{Histogram of detection rate for different event categories. Event categories are ranked based on detection rate when number of proposal is $10,000$ in the ascending order. Top $1-20$, $21-40$, $41-60$ event categories are denote in blue, red, green respectively. Example images for specific event class are shown. Y-axis denotes for detection rate. X-axis denotes for event class name.}}
\vskip -0.5cm
\label{fig:event_hist}
\end{center}
\end{figure*}

\begin{table*}[tp]
\centering
\footnotesize                                                                                          \caption{Full name of event categories.}
\setlength{\tabcolsep}{1.9pt}                                                                                                                                    
\begin{tabulary}{\linewidth}{lcccccccccccccccccccc}                                                                                                                                                   \toprule[1pt]                                                                                                                             
& Traf. & Fest. & Para. & Demo. & Cere. & March. & Bask. & Shop. & Mata. & Acci. & Elec. & Conc. & Awar. & Picn. & Riot. & Fune. & Chee. & Firi. & Raci. & Vote. \\     \midrule                                                                                                                                                              
 &\rotatebox{90}{Traffic} &\rotatebox{90}{Festival} &\rotatebox{90}{Parade} &\rotatebox{90}{Demonstration} &\rotatebox{90}{Ceremony} &\rotatebox{90}{People Marching} &\rotatebox{90}{Basketball} &\rotatebox{90}{Shoppers} &\rotatebox{90}{Matador Bullfighter} &\rotatebox{90}{Car Accident} &\rotatebox{90}{Election Campain} &\rotatebox{90}{Concerts} &\rotatebox{90}{Award Ceremony} &\rotatebox{90}{Picnic} &\rotatebox{90}{Riot} &\rotatebox{90}{Funeral} &\rotatebox{90}{Cheering} &\rotatebox{90}{Soldier Firing} &\rotatebox{90}{Car Racing} &\rotatebox{90}{Voter}\\   
\bottomrule
\toprule                                                                                                                               
& Stoc. & Hock. & Stud. & Skat. & Gree. & Foot. & Runn. & Driv. & Dril. & Phot. & Spor. & Grou. & Cele. & Socc. & Inte. & Raid. & Base. & Patr. & Angl. & Resc. \\
 \midrule                                                                                            
 &\rotatebox{90}{Stock Market} &\rotatebox{90}{Hockey} &\rotatebox{90}{Students Schoolkids} &\rotatebox{90}{Ice Skating} &\rotatebox{90}{Greeting} &\rotatebox{90}{Football} &\rotatebox{90}{Running} &\rotatebox{90}{People Driving Car} &\rotatebox{90}{Soldier Drilling} &\rotatebox{90}{Photographers} &\rotatebox{90}{Sports Fan} &\rotatebox{90}{Group} &\rotatebox{90}{Celebration Or Party} &\rotatebox{90}{Soccer} &\rotatebox{90}{Interview} &\rotatebox{90}{Raid} &\rotatebox{90}{Baseball} &\rotatebox{90}{Soldier Patrol} &\rotatebox{90}{Angler} &\rotatebox{90}{Rescue}\\                                                             

\bottomrule
\toprule                                                                                                         
& Gymn. & Hand. & Wait. & Pres. & Work. & Parach. & Coac. & Meet. & Aero. & Boat. & Danc. & Swim. & Fami. & Ball. & Dres. & Coup. & Jock. & Tenn. & Spa. & Surg.\\       \midrule                                                                                                                                                                                  
 &\rotatebox{90}{Gymnastics} &\rotatebox{90}{Handshaking} &\rotatebox{90}{Waiter Waitress} &\rotatebox{90}{Press Conference} &\rotatebox{90}{Worker Laborer} &\rotatebox{90}{Parachutist Paratrooper} &\rotatebox{90}{Sports Coach Trainer} &\rotatebox{90}{Meeting} &\rotatebox{90}{Aerobics} &\rotatebox{90}{Row Boat} &\rotatebox{90}{Dancing} &\rotatebox{90}{Swimming} &\rotatebox{90}{Family Group} &\rotatebox{90}{Balloonist} &\rotatebox{90}{Dresses} &\rotatebox{90}{Couple} &\rotatebox{90}{Jockey} &\rotatebox{90}{Tennis} &\rotatebox{90}{Spa} &\rotatebox{90}{Surgeons}\\
\bottomrule[1pt]

\end{tabulary}                                                                                                                                                                                                                                                                                     
\caption{Full name of abbreviation of event categories.}                                                                                                                                                                                                                                                                          
\label{table:class_name}                                                                                                                                                                                      \vspace{-8pt}                               
\end{table*}
 
{\small
\bibliographystyle{ieee}
\bibliography{wider_face}
}

\end{document}